\documentclass[]{article}
\usepackage{amsmath}
\usepackage{amssymb}
\usepackage{amsthm} 
\usepackage{cite}
\usepackage{enumitem}
\usepackage{algorithm}
\usepackage{algpseudocode}
\newtheorem{definition}{Definition}[section]
\usepackage{graphicx}
\usepackage{multirow}
\usepackage{booktabs} 
\usepackage{subcaption}
\graphicspath{{images/}{figures/}} 
\usepackage{booktabs}
\usepackage{siunitx}
\usepackage{caption}
\usepackage{geometry}
\usepackage{cleveref}
\usepackage{makecell} 
\usepackage{array}
\usepackage{bm}
\usepackage{authblk}
\newcolumntype{C}{>{\centering\arraybackslash}p{1.8cm}}
\title{MTL-FNO: A Lightweight Multi-Task Fourier Neural Operator for Sparse Field Reconstruction}

\author[1,2]{Siyu Ye}
\author[1,2]{Shihang Li}
\author[1,2]{Zhiqiang Gong\thanks{E-mail address: \texttt{gongzhiqiang13@nudt.edu.cn}}}
\author[1,2]{Benrong Zhang}
\author[1,2]{Weien Zhou}
\author[1]{Yiyong Huang}
\author[1,2]{Wen Yao\thanks{E-mail address: \texttt{wendy0782@126.com}}}

\affil[1]{Defense Innovation Institute, Academy of Military Science, Beijing, 100071, China}
\affil[2]{Intelligent Game and Decision Laboratory, Beijing, 100071, China}

\date{}

\begin{document}

\maketitle

\begin{abstract}
Efficient multi-field sparse reconstruction onboard is essential for the autonomous operation of aerospace vehicles. While existing deep learning models exhibit promise for single-field reconstruction, deploying multiple independent models leads to prohibitive model size growth and fails to exploit cross-field correlations, particularly under few-shot conditions. To address these challenges, we first propose a lightweight multi-task Fourier neural operator (MTL‑FNO), an end‑to‑end joint training framework based on hard parameter sharing. In each MTL-FNO layer, the parameters are divided into shared and task-specific components to capture common features across fields while preserving task-specific characteristics. Moreover, the task-specific fine-tuning parameters are implemented as low-rank terms, achieving substantial model compression. Second, to address the difficulty of co-optimizing shared and task-specific fine-tuning parameters together with their real and imaginary components, we revisit the FNO's spectral weight from a polar-form perspective and devise a physically meaningful decoupled optimization scheme. Specifically, we apply the polar decomposition to slice-wise disentangle the spectral weight into a unitary tensor encoding phase information and a positive semi-definite Hermitian tensor characterizing amplitude. By decoupling the optimization of phase and amplitude, our method can effectively mitigate the task conflicts. Meanwhile, to preserve unitary geometric fidelity during training, the Cayley transform is introduced to reparameterize the unitary tensor, relaxing the constrained optimization on the unitary manifold to an unconstrained one. Finally, the effectiveness of the proposed method under few-shot conditions is validated on two representative engineering cases. Results show that MTL‑FNO achieves reconstruction accuracy comparable to or even surpassing that of vanilla FNO, while reducing the total model size by 76\% and 60\% in the two cases, respectively.

\textbf{Keywords}: Lightweight, Multi-Task FNO, Sparse Field Reconstruction, Polar Decomposition, Reparameterization.
\end{abstract}

\section{Introduction}

The autonomous operation of next-generation aerospace vehicles relies heavily on the capability of self-state perception\cite{Wang2025Spacecraft}. A critical aspect of this capability is to promptly and accurately reconstruct multiple coupled physical fields or a specific field under varying operating conditions from sparse sensor observations\cite{fukami2021gfr}. To this end, deep learning has recently emerged as a powerful paradigm, leveraging its exceptional nonlinear approximation and high-dimensional feature extraction capabilities\cite{Sharma2018Heat,Marco2022GNN}. Conventionally, raw sensor data are transmitted to Earth for back-end processing, incurring significant latency and bandwidth overhead\cite{Manoj2023onboard}. This has motivated a growing trend toward deploying deep learning models onboard for data processing in situ\cite{Lorenzo2024Onboard}. However, unlike ground-based systems, onboard computing platforms are severely constrained in terms of computational power, memory capacity, and power budgets. These constraints are further exacerbated when simultaneous multi-field reconstruction is required. Consequently, striking an optimal trade-off between prediction accuracy and computational complexity within a multi-field context has emerged as the central bottleneck for deploying deep learning models onboard. To overcome this bottleneck, two major challenges must be addressed: (i) identifying a deep learning model framework that delivers superior computational efficiency and prediction accuracy for single-field reconstruction; and (ii) extending this framework to multi-field reconstruction in a manner that effectively exploits cross-field commonalities while preserving high parameter efficiency.

To meet the first challenge, a variety of models have demonstrated great promise in single-field reconstruction, notably convolutional neural networks (CNNs) based on the U-Net architecture\cite{Olaf2015U-Net,Zhao2023cnn,Jordan2024U-Net,Tong2025PIU-Net}, neural operators including DeepONet\cite{Lu2021DeepONet} and FNO\cite{Li2021FNO}, Transformer-based methods such as Senseiver\cite{Santos2023Senseiver}, and generative models like PhySense\cite{Ma2025PhySense}. Despite their effectiveness, many of these approaches rely on deeply stacked architectures, complex attention-based mechanisms or intricate generative sampling procedures, resulting in large model footprints and limited computational efficiency. Their deployment on onboard devices typically necessitates additional model compression or acceleration techniques\cite{Zhu2025pruning, Babak202Quantization, Amir2025Distillation, OU2024lowrank}. To date, while numerous models have been specifically developed for spaceborne or airborne applications—primarily targeting vision-centric tasks such as image segmentation, classification\cite{Jon2025On1D-CNN, Roberto2025onnas}, and object detection\cite{Dimitrios2025Edge, Bharadwaj2026onboard}—dedicated solutions for physical field reconstruction remain comparatively scarce. Among the aforementioned reconstruction models, FNO leverages a low-frequency truncation strategy to effectively control both parameter count and computational cost. More importantly, by performing global convolutions in the frequency domain via the Fast Fourier Transform (FFT), FNO efficiently captures long-range spatial dependencies and enables holistic information propagation. As demonstrated in \cite{Sarthak2022Comparison}, FNO has achieved a favorable balance between accuracy and efficiency across many partial differential equation (PDE) learning tasks\cite{Sarthak2022Comparison}, making it a promising lightweight model framework for efficient single‑field reconstruction.
 
The second challenge—extending this single‑field framework to multi‑field reconstruction while preserving parameter efficiency—rules out obvious but impractical solutions. For instance, deploying multiple independent models onboard, each dedicated to one reconstruction task, faces two critical drawbacks. First, the total parameter count and storage footprint scale linearly with the number of tasks, quickly exceeding the resource limits of onboard hardware. Second, training a separate model for each new task requires substantial amounts of high-fidelity data\cite{Jinghong2025lamdaFNO,Yu2024TransferFNO}, but acquiring such data for aerospace physical fields is costly\cite{Brunton2021Data-Driven}. This makes multiple independent modeling somewhat impractical. Ma et al.\cite{Ma2025multisource} pointed out that under similar geometric configurations, boundary conditions and external excitation, multiple physical fields are correlated. Building on this insight, they proposed a reduced‑order surrogate model that extracts coherent structures from multiple fields, thereby decreasing the number of sensors while improving reconstruction accuracy. However, their methods still perform independent dimensionality reduction and separate modeling for each field, resulting in suboptimal parameter efficiency. These limitations strongly motivate the adoption of hard-parameter sharing based multi-task learning (MTL) framework\cite{Maxime2024Multitask}. Hard-parameter sharing is a classic MTL paradigm, wherein multiple tasks share common hidden layers to learn a joint representation of generic features. By virtue of this design, the model can extract common latent features from multiple fields—thus reducing the reliance on high‑fidelity samples, while consolidating multiple independent models into a compact architecture, drastically compressing the overall parameter count.

Many hard parameter sharing approaches\cite{Wang2023AdaptiveHPS,Graham2023Mtl,Zhang2023MTL,Cheng2025Multitask} typically employ task-specific output heads while sharing the majority of hidden layers across tasks. However, such rigid parameter sharing often suppresses or overlooks task-specific characteristics, potentially leading to negative transfer. Recently, Garg et al.\cite{garg2025ftn} proposed the Factorized Tensor Network (FTN), a parameter-efficient fine-tuning (PEFT) strategy that introduces low-rank, task or domain specific tensor on top of a frozen pretrained backbone. FTN achieves performance comparable to those independent single task or domain models while adding only a minimal number of trainable parameters. Similar PEFT techniques have been widely adopted in large language models and other domain\cite{Wang2023PEM,Zhang2025MoRE,He2025RaSA,Yang2024MTL-LoRA}. Nevertheless, these PEFT strategies critically depend on the availability of a high-quality pretrained backbone, a condition that is readily met in data-rich fields. In contrast, for physical field reconstruction, no universal pretrained backbone exists, making it infeasible to adopt the conventional paradigm of freezing a shared backbone and training only task-specific modules. Moreover, compared with parameter optimization in the real‑valued domain, complex‑valued spectral operations in FNOs—which must simultaneously account for the aggregation of shared and task‑specific components, as well as the coordinated optimization of real and imaginary parts—are more susceptible to task conflicts, which would degrade the performance of multi‑task models. To address these issues, on the one hand, we apply an end‑to‑end joint training framework that does not freeze the backbone network; both the shared backbone and the task‑specific low‑rank adaptation modules are optimized simultaneously. On the other hand, from a polar-form perspective on the complex-valued spectral weights in FNOs, the matrix slice at each frequency mode can be interpreted as amplitude scaling and phase mixing across channels. Optimizing the real and imaginary parts of complex-valued parameters is, in essence, adjusting the coupled phase–amplitude configuration. Therefore, we adopt a decoupling strategy to alleviate the optimization difficulty induced by the coupling within the frequency domain.

In this work, we propose a lightweight multi-task Fourier Neural Operator (MTL-FNO), which adopts layer-wise partitioning of shared and task-specific parameters as well as an end-to-end joint training paradigm. In each MTL-FNO layer, task‑specific fine‑tuning parameters are implemented as lightweight terms constructed via CANDECOMP/PARAFAC (CP) tensor decomposition, and added separately to the shared parameters. Meanwhile, polar decomposition and Cayley transform are applied. The complex‑valued weight tensor is decoupled into a unitary tensor encoding phase information and a positive semi‑definite Hermitian tensor characterizing amplitude, after which the constrained optimization over the unitary manifold is relaxed into an unconstrained one. Overall, our contributions can be summarized as follows:

\begin{itemize}[label=\textbullet, left=0pt, labelwidth=1.5em, labelsep=0.9em, align=left, itemindent=!]
	\item We propose MTL-FNO, an end-to-end joint training framework based on hard parameter sharing. By decomposing each layer’s parameters into shared and task-specific components and employing low-rank terms as task-specific fine-tuning parameters, the model effectively captures cross-field commonalities while preserving task-specific characteristics, thereby achieving substantial model compression.

    \item We revisit FNO's spectral convolution from a polar-form perspective, and concurrently employ polar decomposition and the Cayley transform to decouple the optimization of phase and amplitude in frequency domain, thereby mitigating task conflicts and enhancing the collaborative representational capacity of the multi-task model. 

    \item Extensive experiments on two representative engineering datasets demonstrate that MTL-FNO outperforms state-of-the-art baselines in predicted accuracy, and simultaneously achieving significant model compression under few-shot conditions.
\end{itemize}

The remainder of this paper is organized as follows: \Cref{sec:rw} reviews related work on Fourier Neural Operator, Factorized Tensor Network, as well as the definitions of polar decomposition and the Cayley transform. \Cref{sec:method} presents the proposed MTL-FNO in detail. \Cref{sec:experiment} evaluates the method through two representative engineering case, one is the joint reconstruction of temperature, pressure, and stress fields in a 2D blunt-wedge hyper-sonic flow problem, and the other is temperature field reconstruction inside a satellite cabin under varying conditions. Finally, \cref{sec:conclu} concludes the paper and outlines promising directions for future research.

\section{Preliminaries}
\label{sec:rw}

\textbf{Fourier Neural Operator:} Neural operators extend conventional neural networks by learning mappings between infinite-dimensional function spaces, rather than point-to-point transformations in finite-dimensional vector spaces. FNO leverages the FFT to map input features from the spatial domain into the frequency domain, where global spectral convolutions are performed to effectively capture long-range dependencies inherent in physical fields. As illustrated in Fig.\ref{fig:MTL-FNO}(b), the output of the $l$-th Fourier layer can be expressed as 

\begin{equation}
v^l = \sigma\left( W^l v^{(l-1)} + \mathcal{F}^{-1}\left( \mathcal{R}^l \odot \mathcal{F}(v^{(l-1)}) \right) + b^l \right),
\end{equation}
where $\sigma:\mathbb{R} \to \mathbb{R}$ denotes a point-wise nonlinear activation, $\mathcal{F}$ and $\mathcal{F}^{-1}$ are the Fourier and inverse Fourier transforms. $\mathcal{R}^l$ is a learnable complex-valued weight tensor. $W^l$ and $b^l$ represent a spatial-domain linear transform and bias, and $\odot$ refers to slice‑wise matrix multiplication. For an input discretized over $m$ grid points, standard convolutions incur $\mathcal{O}(m^2)$ complexity, whereas FFT-based global convolution reduces it to $\mathcal{O}(m \log m)$. Moreover, by exploiting the rapid decay of Fourier coefficients—most physical fields concentrate energy in low frequencies—only the first $k$ modes ($k \ll m$) are retained, so that $\mathcal{R}^l \in \mathbb{C}^{k \times k \times C \times C}$ and its parameter count scale as $\mathcal{O}(k^2\cdot C^2)$. This spectral truncation yields substantial model compression with negligible accuracy loss, thus striking a favorable balance between model size and accuracy.

\textbf{Factorized Tensor Network:} Garg et al.\cite{garg2025ftn} proposed the Factorized Tensor Network (FTN), a parameter-efficient fine-tuning strategy designed for multi-task and multi-domain learning. The central idea of FTN is to augment a shared frozen backbone, inherited from a source model, with task- or domain-specific low-rank tensor factors. Formally, for the $l$-th layer and the $i$-th task, the adapted weight $W_i^l$ is composed of a shared weight $W_{\text{share}}^l$ and a task-specific fine-tuning term $\Delta W_i^l$, which can be formulated as
\begin{equation}
	\begin{aligned}
		& W_i^l = W_{\text{share}}^l + \Delta W_i^l, \\ 
		&\Delta W_i^l = \sum_{r=1}^{R} \lambda_r \mathbf{a}_{i,r}^{l,(1)} \circ \mathbf{a}_{i,r}^{l,(2)} \circ \mathbf{a}_{i,r}^{l,(3)},
	\end{aligned}
	\label{eq:cp}
\end{equation}
where $\circ$ denotes the outer product, $\mathbf{a}_{i,r}^{l,(j)}$ are task-specific factors along the $j$-th mode, and $\lambda_r$ is a scaling coefficient, $R$ is the CP decomposition rank. The shared backbone $W_{\text{share}}^l$ remains frozen during training, while only the low-rank factors $\mathbf{a}_{i,r}^{l,(j)}$ are updated per task. This factorization allows each task to introduce only a minimal number of additional parameters—scaling with the rank and the lengths of the factor vectors rather than with the full weight tensor size—while maintaining accuracy on par with fully independent single‑task or single‑domain networks. 

\begin{definition}[Polar Decomposition]\label{def:polar}
	For any non-singular square matrix $H \in \mathbb{C}^{k \times k}$, the polar decomposition states that $H$ can be uniquely factorized as
	\begin{equation}
		\begin{aligned}
			&H = U \cdot P, \\
			&U^\dagger U = I, \\
			&P = P^\dagger, \quad \lambda(P) \geq 0,
		\end{aligned}
		\label{eq:polar}
	\end{equation}
	where $U$ is a unitary matrix, $P$ is a positive semi-definite Hermitian matrix, $\lambda(P)$ denotes the eigenvalues of $P$, and $\dagger$ stands for the conjugate transpose. In the context of FNO spectral weights, this decomposition allows us to decouple the phase shift ($H$) from the amplitude scaling ($P$). 
\end{definition}

\begin{definition}[Cayley Transform]\label{def:cayley} 
	For any skew-Hermitian matrix $K$ (i.e., $K^\dagger=-K$) such that $I+K$ is invertible, the Cayley transform is defined as
	\begin{equation}
		U = (I - K)(I + K)^{-1}.
		\label{eq:cayley}
	\end{equation}
	The resulting $U$ is guaranteed to be unitary. Moreover, the mapping is smooth and differentiable, allowing unconstrained optimization over $K$ to serve as a surrogate for constrained optimization over the unitary manifold.
\end{definition}

\section{Methodology: The MTL-FNO Framework}
\label{sec:method}

\textbf{Problem Setup:} Consider $T$ independent reconstruction tasks, each associated with sparse sensor observation data $\{\mathbf{s}_1, \mathbf{s}_2, \cdots, \mathbf{s}_T \}$, where $\mathbf{s}_i \in \mathbb{R}^n$ denotes the observations from $n$ sensors in the $i$-th task. Correspondingly, the global physical fields are denoted as $\{ \mathbf{Y}_1, \mathbf{Y}_2, \cdots, \mathbf{Y}_T \}$, with $\mathbf{Y}_i \in \mathbb{R}^m$ representing the field values at $m$ discrete spatial locations ($n \ll m$). The objective is to learn a unified multi-task model parameterized by $\Theta$, which represents the full set of trainable shared and task-specific parameters. Such that for any task $i$, the model realizes a mapping $G(\cdot; \Theta):\mathbb{R}^{n} \to \mathbb{R}^m$ and predicts a field $\hat{\mathbf{Y}}_i = G(\mathbf{s}_i;\Theta)$ that closely approximates the ground-truth $\mathbf{Y}_i$. 

\subsection{Model architecture}

\begin{figure}[htbp]
	\centering
	\includegraphics[width=0.85\linewidth]{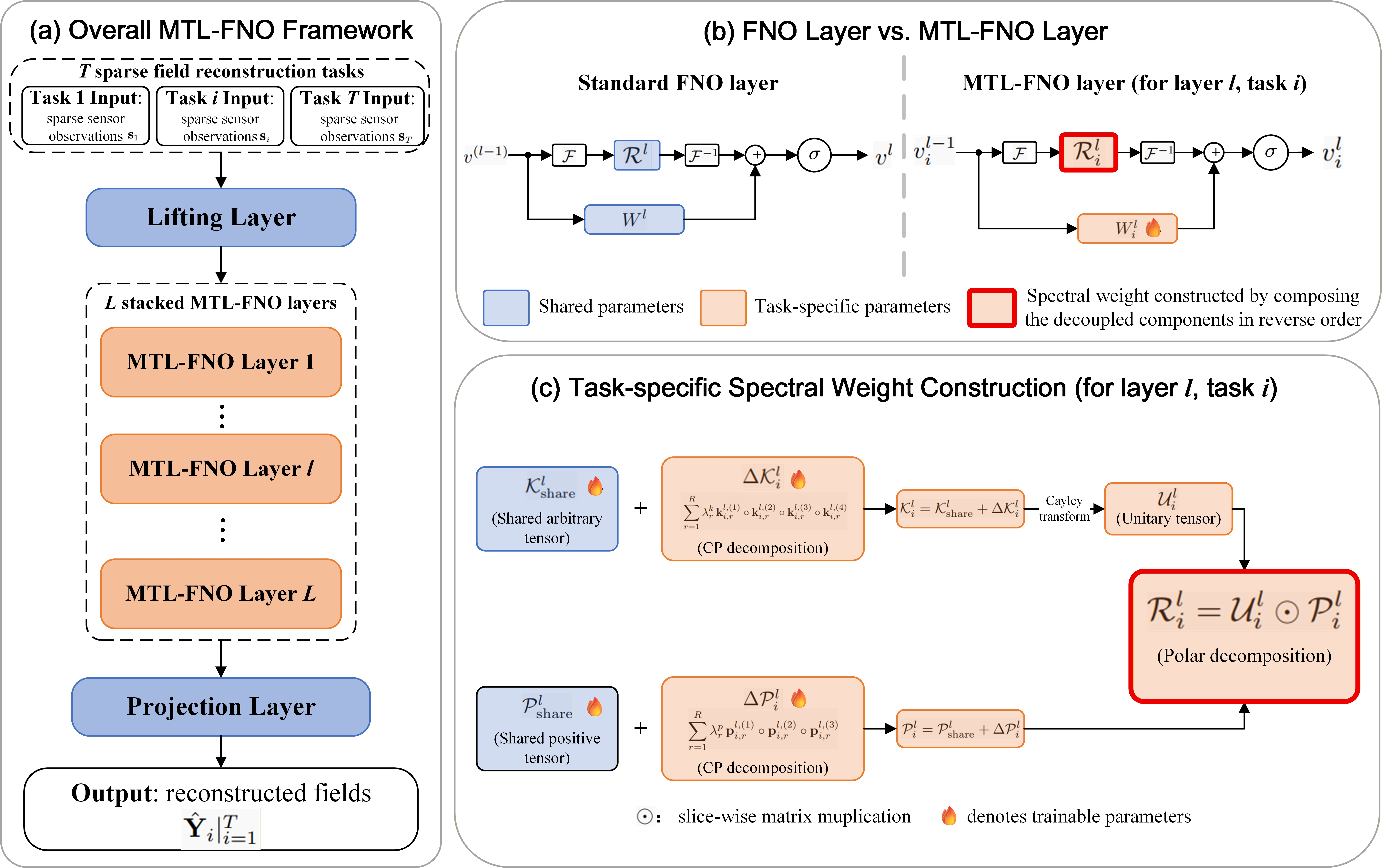}
	\caption{(a) Overall Framework of the proposed MTL-FNO. (b) Comparisons between the standard FNO layer and MTL-FNO layer. (c) Schematic illustration of the final spectral weight construction.}
	\label{fig:MTL-FNO}
\end{figure}

The overall architecture of the proposed MTL-FNO is depicted in Fig.\ref{fig:MTL-FNO}(a). Given the model consists of $L$ stacked layers, parameters in each layer are partitioned to shared and task-specific trainable components, and the final spectral weight $\mathcal{R}_i^l$ corresponding to the $i$-th task is generated through a series of transformations, optimizations, and aggregations applied to these two components. It is worth noting that except $\mathcal{R}_i^l$, spatial convolution weights $W_i^{l}$ and biases are task-specific, while the lifting and projection layers are shared across all tasks. For the $i$-th task, the output $v_i^l$ of MTL-FNO at the $l$-th layer can be expressed as:
\begin{equation}
	v_i^l = \sigma\Big( W_i^l v_i^{l-1} + \mathcal{F}^{-1}\big( \mathcal{R}_i^l \odot \mathcal{F}(v_i^{l-1}) \big) + b_i^l \Big). 
	\label{eq:MTL-FNO}
\end{equation}

The core design of MTL-FNO lies in the decoupling and task-adaptive fine-tuning of the complex-valued spectral weights within each layer. In practice, the final weight $\mathcal{R}_i^l$ is reconstructed by composing the decoupled components in reverse order, as illustrated in Fig.\ref{fig:MTL-FNO}(c). The procedure primarily consists of the following steps: (i) computing the task-specific fine-tuning term based on CP tensor decomposition, and added separately to the shared parameters; (ii) obtaining the unitary component via the Cayley transform; and (iii) generating the final task-specific spectral weight by combining the unitary tensor and the positive semi-definite tensor via polar decomposition. In the following, each of these steps is described in detail. 

\subsection{Decoupling Spectral Weight from Polar-Form Perspective}

Recall from \cite{Li2021FNO} that in the vanilla FNO, the complex-valued spectral weight serves as a spectral filter. The matrix slice associated with each frequency mode $(\xi_i,\xi_j)$ comprises complex-valued elements, which can be expressed as $r = A(\cdot)e^{i \phi( \cdot)}$ in polar form. As illustrated in Fig.\ref{fig:cpandcayley}(a), the magnitude $A(\cdot)$ governs the amplitude scaling across channels, whereas $\phi(\cdot)$ controls the phase shifts between channels. These two components enable adaptive scaling and phase rotation of each frequency mode $(\xi_i,\xi_j)$, thereby endowing the model with the capacity to extract and recombine multi-scale physical features in the frequency domain. In the hard parameter sharing based multi-task framework considered in this paper, directly applying a task-specific fine-tuning term $\Delta \mathcal{R}_i^l$ to the shared spectral weight $\mathcal{R}_{\text{share}}^l$ requires the joint optimization and subsequent superposition of their real and imaginary parts. Owing to the coupled nature of these complex-valued parts, this inevitably introduces optimization difficulty and can easily lead to task interference. For example, in polar form, one task may require amplification of a specific frequency mode, while another task may only need to adjust its propagation direction. Such conflicting demands are prone to induce negative transfer, undermining the generalization benefits of multi-task learning (see \cref{subsubsec:ablation} for a detailed ablation study). 

To address this, instead of directly adding the task-specific $\Delta \mathcal{R}_i^l$ on the shared $\mathcal{R}_{\text{share}}^l$, this paper employs a slice-wise polar decomposition to the original spectral weight. The decomposed components $\mathcal{U}_i^l$ and $\mathcal{P}_i^l$ are then obtained via Eq.~\eqref{eq:polar}:

\begin{equation}
	\mathcal{R}_i^l = \mathcal{U}_i^l \odot \mathcal{P}_i^l.
	\label{eq:MTL-FNOpolar}
\end{equation}
where $\mathcal{U}_i^l$ denotes a unitary tensor whose matrix slices are unitary, and $\mathcal{P}_i^l$ is a positive semi-definite tensor. Together, these two tensors replace the original spectral weights and serve as the subjects for layer-wise partitioning and task-adaptive fine-tuning. To balance expressiveness and efficiency, each matrix slice of $\mathcal{P}_i^l$ is assumed to be diagonal and represented compactly as a real-valued vector, so that $\mathcal{P}_i^l \in \mathbb{R}^{k \times k \times C}$. During matrix multiplication with slices of $\mathcal{U}_i^l \in \mathbb{C}^{k \times k \times C \times C}$, these real-valued vectors are broadcast along the channel dimension.

Within the FNO-based framework, this polar decomposition carries clear physical interpretation. Each matrix slice of the unitary tensor, denoted as $U_{\xi_i,\xi_j}^l$, redistributes phase relationships across channels without altering total signal energy, i.e., $\| U_{\xi_i,\xi_j}^l \cdot \textbf{x} \|_2 = \| \textbf{x} \|_2$ for any vector $\textbf{x}$, analogous to lossless wave propagation in conservative systems. Conversely, each slice of the positive tensor independently scales the energy gain per channel without affecting phase structure, mirroring amplitude scaling in active or passive media. By explicitly decoupling phase and amplitude optimization, our approach enables independent and task-adaptive fine-tuning of spectral responses in multi-task scenarios, thereby enhancing collaborative representational capability of model. It should be emphasized that this physical interpretation primarily serves to explain the internal mechanism of spectral mapping in FNO, and does not imply that the entire network strictly adheres to conservation or dissipation laws of real physical systems. Rather, the decomposition provides a principled and interpretable inductive bias that improves the flexibility and robustness of the multi-task sharing architecture.

\subsection{Reparameterizing unitary tensor via Cayley Transform}

\begin{figure}[htbp]
	\centering
	\includegraphics[width=0.85\linewidth]{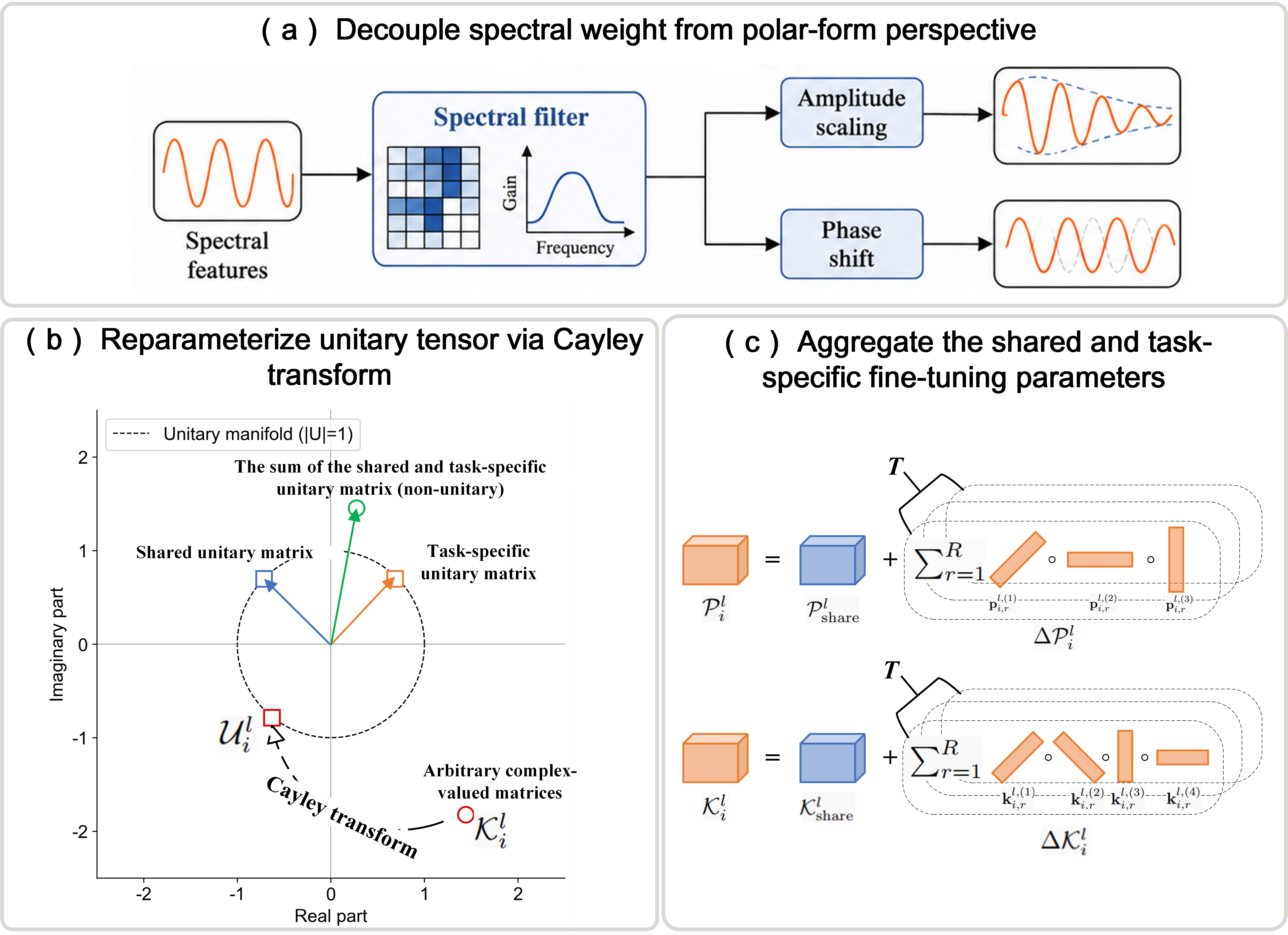}
	\caption{(a) From a polar-form perspective, the complex-valued spectral weight acts as a spectral filter, whose elements govern the amplitude scaling and the phase shifts across channels. (b) The sum of the shared and task-specific unitary matrix is non-unitary; nevertheless, any arbitrary complex-valued matrix can be mapped onto the unitary manifold via the Cayley transform. (c) The task-specific fine-tuning parameters are constructed via CP tensor decomposition, and added separately to the shared parameters.}
	\label{fig:cpandcayley}
\end{figure}

According to the definition of polar decomposition, the unitary matrix must strictly preserve its unitary property, which poses a severe challenge for model training. A natural strategy is to constrain the optimization onto the unitary manifold. However, in our framework, the shared and task-specific fine-tuning parameters need to be co-optimized and subsequently aggregated; even if these two parameters are individually constrained to be unitary, Their sum are non-unitary, as illustrated in Fig.\ref{fig:cpandcayley}(b). Moreover, without imposing any constraints, directly optimizing the shared and task-specific fine-tuning parameters would allow the converged solutions to be arbitrary complex-valued matrices, thereby losing their unitary property. These compromises the desired separation between phase and amplitude optimization, and consequently degrade model performance (see \cref{subsubsec:ablation} for a detailed ablation study). 

To preserve unitary geometric fidelity while maintaining differentiability, we apply the Cayley transform introduced in \cref{sec:rw} to reparameterize the unitary tensor $\mathcal{U}_i^l$ slice-wise, thereby relaxing the optimization over the unitary manifold to an unconstrained optimization over the arbitrary complex-valued matrices space. Notably, in the implementation, the skew-Hermitian component $\hat{\mathcal{K}}_i^l$ is first extracted via anti-symmetrization, and then the Cayley transform is performed. The procedure can be expressed as
\begin{equation}
	\begin{aligned}
		\mathcal{U}_i^l &= (I - \hat{\mathcal{K}}_i^l)(I + \hat{\mathcal{K}}_i^l)^{-1}, \\
		\hat{\mathcal{K}}_i^l &= \tfrac{1}{2}\big( \mathcal{K}_i^l - (\mathcal{K}_i^l)^\dagger \big).
	\end{aligned}
	\label{eq:MTL-FNOpolar}
\end{equation}
Therefore, the arbitrary component $\mathcal{K}_i^l$ is employed to substitute $\mathcal{U}_i^l$ and is subjected to layer-wise partitioning into shared and task-specific parameters. In summary, as illustrated in Fig.\ref{fig:cpandcayley}(c), the shared trainable parameters of MTL-FNO comprise a complex-valued tensor $\mathcal{K}_{\text{share}}^{l}$ and a positive semi-definite tensor $\mathcal{P}_{\text{share}}^{l}$. The task-specific fine-tuning parameters consist of complex-valued factors $\mathbf{k}^{l,(1)}_{i,r}, \mathbf{k}^{l,(2)}_{i,r}, \mathbf{k}^{l,(3)}_{i,r},\mathbf{k}^{l,(4)}_{i,r}$ and real-valued factors $\mathbf{p}^{l,(1)}_{i,r}, \mathbf{p}^{l,(2)}_{i,r},\mathbf{p}^{l,(3)}_{i,r}$. Through CP tensor decomposition introduced in Eq.\eqref{eq:cp}, $\Delta \mathcal{K}_i^{l} \in \mathbb{C}^{k \times k \times C \times C}$ and $\Delta \mathcal{P}_i^{l} \in \mathbb{R}^{k \times k \times C}$ can be constructed: 
\begin{equation}
	\begin{aligned}
		&\Delta \mathcal{K}_i^{l} = \sum_{r=1}^{R} \lambda_r^k \, \mathbf{k}^{l,(1)}_{i,r} \circ \mathbf{k}^{l,(2)}_{i,r} \circ \mathbf{k}^{l,(3)}_{i,r} \circ \mathbf{k}^{l,(4)}_{i,r}, \\ 
		&\Delta \mathcal{P}_i^l = \sum_{r=1}^{R} \lambda_r^p \, \mathbf{p}^{l,(1)}_{i,r} \circ \mathbf{p}^{l,(2)}_{i,r} \circ \mathbf{p}^{l,(3)}_{i,r},
	\end{aligned}
	\label{eqcp}
\end{equation}
where $\mathbf{k}^{l,(j)}_{i,r},\mathbf{p}^{l,(j)}_{i,r}$ represent the task-specific factor vectors along the $j$-th mode, $\lambda_r^k, \lambda_r^p  \in \mathbb{R}$ are the scaling coefficient. By subsequently aggregating the shared and task‑specific parts, $\mathcal{P}_i^l, \mathcal{K}_i^l$ of the $i$-th task at the $l$-th layer are obtained:
\begin{equation}
	\begin{aligned}
		\mathcal{P}_i^l &= \mathcal{P}_{\text{share}}^l + \Delta \mathcal{P}_i^l, \\
		\mathcal{K}_i^l &= \mathcal{K}_{\text{share}}^l + \Delta \mathcal{K}_i^l.
	\end{aligned}
	\label{eq:MTL-FNOparameters}
\end{equation}

\subsection{Model Training}
The MTL-FNO framework is trained end-to-end by minimizing a composite objective that aggregates the reconstruction errors across all $T$ tasks. The total loss $\mathcal{L}_{total}$ is formulated as a sum of these errors:
\begin{equation}
	\mathcal{L}_{total} = \min_{\Theta} \sum_{i=1}^{T} \| \mathbf{Y}_i - \hat{\mathbf{Y}}_i(\mathbf{s}_i; \Theta) \|_2,
	\label{eq:totalloss}
\end{equation}
To further balance these multi-task objectives, Alignment-MTL is employed \cite{Senushkin2023Alignment}. The complete training procedure of MTL-FNO is summarized in Algorithm~\ref{alg:mtl_fno}. 
 
 \begin{algorithm}[ht]
 	\caption{Training Algorithm for MTL-FNO}
 	\label{alg:mtl_fno}
 	\begin{algorithmic}[1]
 		\State \textbf{Input:} Dataset $\mathcal{D} = \{(\mathbf{s}_i, \mathbf{Y}_i)\}_{i=1}^T$
 		\State \textbf{Output:} A multitask model that realizes mappings $G(\cdot; \Theta)$ for all tasks
 		\State \textbf{Initialization:}
 		\State \quad Shared parameters: $\{\mathcal{K}_{\text{share}}^l, \mathcal{P}_{\text{share}}^{l} \}_{l =1}^L$
 		\State \quad Task-specific parameters: $\{ \mathbf{k}^{l,(1)}_{i,r}, \mathbf{k}^{l,(2)}_{i,r}, \mathbf{k}^{l,(3)}_{i,r},\mathbf{k}^{l,(4)}_{i,r}, \mathbf{p}^{l,(1)}_{i,r}, \mathbf{p}^{l,(2)}_{i,r},\mathbf{p}^{l,(3)}_{i,r},W_i^l\}_{i=1,2,\cdots, T; l=1,2,\cdots,L}$
 		\For{$epoch = 1$ \textbf{to} $N_{epoch}$}
 		\State \textbf{Forward Pass:}
 		\State \quad Compute low-rank fine-tuning terms:
 		\Statex \quad\quad\quad $\Delta \mathcal{K}_i^l \gets \sum_{r=1}^{R} \lambda_r^k \, \mathbf{k}_{i,r}^{l,(1)} \circ \mathbf{k}_{i,r}^{l,(2)} \circ \mathbf{k}_{i,r}^{l,(3)} \circ \mathbf{k}_{i,r}^{l,(4)}$
 		\Statex \quad\quad\quad $\Delta \mathcal{P}_i^l \gets \sum_{r=1}^{R} \lambda_r^p \, \mathbf{p}_{i,r}^{l,(1)} \circ \mathbf{p}_{i,r}^{l,(2)} \circ \mathbf{p}_{i,r}^{l,(3)}$
 		\State \quad Aggregate parameters:
 		\Statex \quad\quad\quad $\mathcal{K}_i^l \gets \mathcal{K}_{\text{share}}^l + \Delta \mathcal{K}_i^l$ \quad 
 		\Statex \quad\quad\quad $\mathcal{P}_i^l \gets \mathcal{P}_{\text{share}}^l + \Delta \mathcal{P}_i^l$ \quad 
 		\State \quad Extract skew-Hermitian component: $\hat{\mathcal{K}}_i^l \gets \frac{1}{2}\left( \mathcal{K}_i^l - (\mathcal{K}_i^l)^\dagger \right)$   \quad \quad \quad   
 		\State \quad Apply Cayley transform: ${\mathcal{U}_i^l} \gets (I - \mathcal{\hat{K}}_i^l)(I + \mathcal{\hat{K}}_i^l)^{-1}$
 		\State \quad Generate final task-specific spectral weights: $\mathcal{R}_i^l \gets \mathcal{U}_i^l \odot \mathcal{P}_i^l$ 
 		\State \quad Compute predictions $\hat{\mathbf{Y}}_i|_{i=1}^T$ and evaluate total loss
 		\State  \textbf{Backward Pass:}
 		\State \quad Jointly update shared and task-specific parameters $\Theta$
 		\EndFor		
 	\end{algorithmic}
 \end{algorithm}

\clearpage
\section{Experiments}
\label{sec:experiment}

\subsection{Experimental Cases}

\subsubsection{Case A: Multi-Physics Field Reconstruction for 2D Blunt-Wedge Rarefied Flow}

This case focuses on the hypersonic rarefied flow over a two-dimensional blunt wedge. The training and test datasets are generated from high-fidelity aerothermodynamic simulations by directly solving the Boltzmann equation using the Unified Gas-Kinetic Scheme. The dataset fully resolves the spatial distributions of temperature T, pressure P and stress components $({\tau_{xx}, \tau_{xy}, \tau_{yy}})$. According to the planar symmetry of the flow field, only the upper half of the computational domain is modeled, comprising $m=3792$ grid points (see Appendix for detailed physical conditions). The model input consists of observations from $n=3$ sensors placed on the wedge surface—mimicking a realistic scenario where limited sensor observations on an aircraft wing are used to reconstruct the full-field physical quantities. The case is designed to evaluate MTL-FNO’s capability for joint multi-physics modeling under the few-shot conditions.

\begin{figure}[htbp]
	\centering
	\includegraphics[width=0.3\linewidth]{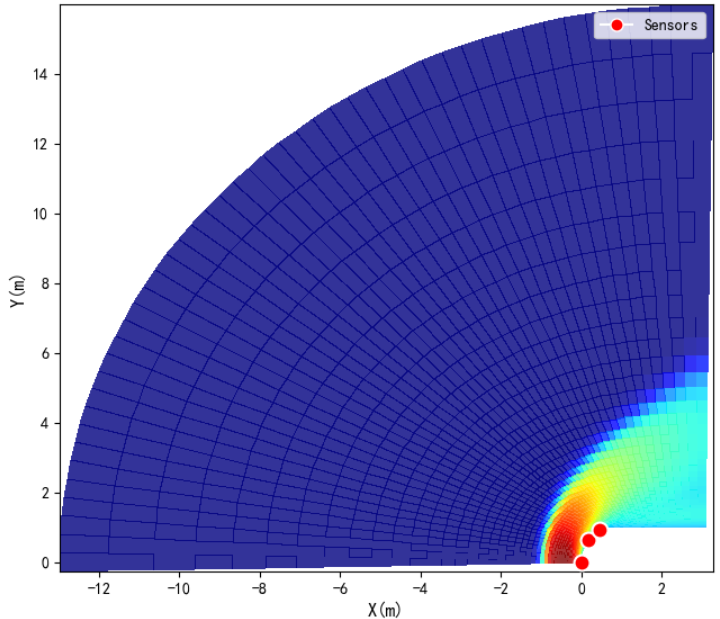}
	\caption{Sparse sensor array configurations on the blunt wedge.}
	\label{fig:dunxiesensorplacement}
\end{figure}

\subsubsection{Case B: Multi-condition Temperature Field Reconstruction for Satellite Cabin}
This case adopts the publicly available benchmark dataset introduced by Chen et al.\cite{Chen2023benchmark} to assess model generalization across diverse thermal boundary conditions. The dataset encompasses three representative in-orbit thermal environments for satellites: fully Dirichlet boundary conditions with active cooling (ADlet), dynamic sinusoidal thermal perturbations (DSine), and high heat-sink conditions (HSink). The mesh configuration, boundary setups, and sensor placements strictly follow the specifications in\cite{Chen2023benchmark}. The input of dataset comprises observations from $n=32$ sparse sensors, while the output is the full-field temperature distribution over $m=40000$ grid points inside the satellite cabin. By performing multi-task modeling on the same physical field under diverse conditions, the adaptability of MTL-FNO can be validated, thereby providing a potential for its application in complex spacecraft thermal control systems.

\begin{figure}[htbp]
	\centering
	\begin{subfigure}[b]{0.3\textwidth}
		\centering
		\includegraphics[width=\linewidth]{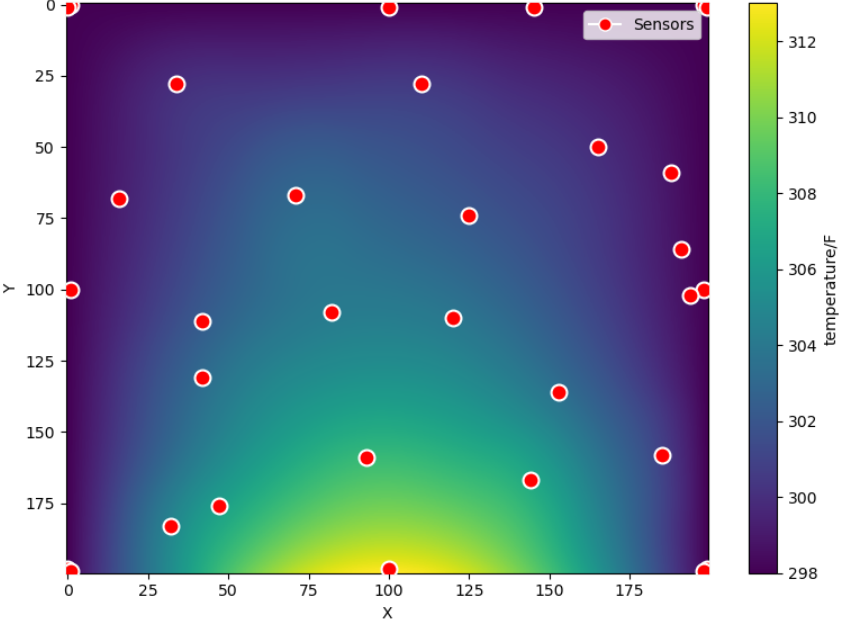}
		\caption{ADlet condition}
		\label{fig:ADletsensors}
	\end{subfigure}
	\hfill
	\begin{subfigure}[b]{0.3\textwidth}
		\centering
		\includegraphics[width=\linewidth]{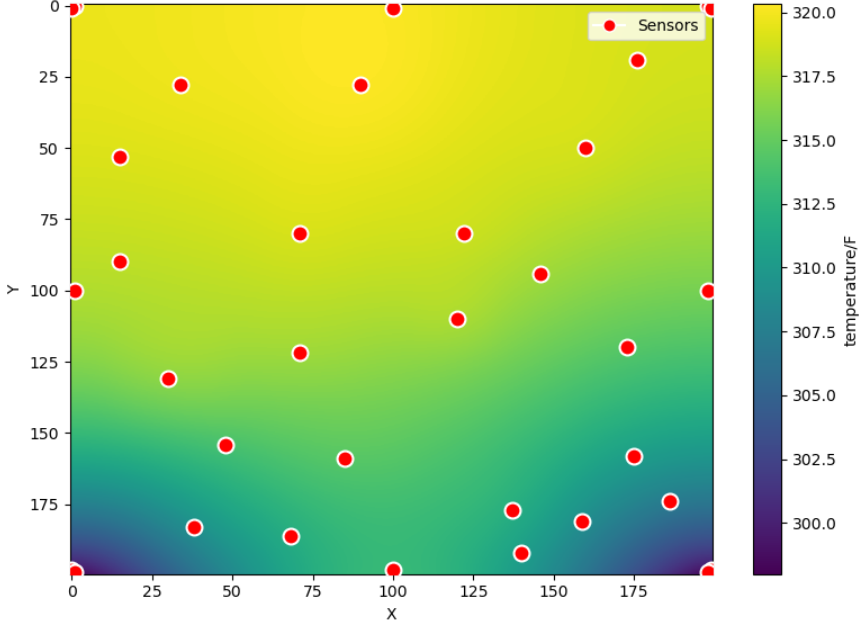}
		\caption{DSine condition}
		\label{fig:Dsinesensors}
	\end{subfigure}
	\hfill
	\begin{subfigure}[b]{0.3\textwidth}
		\centering
		\includegraphics[width=\linewidth]{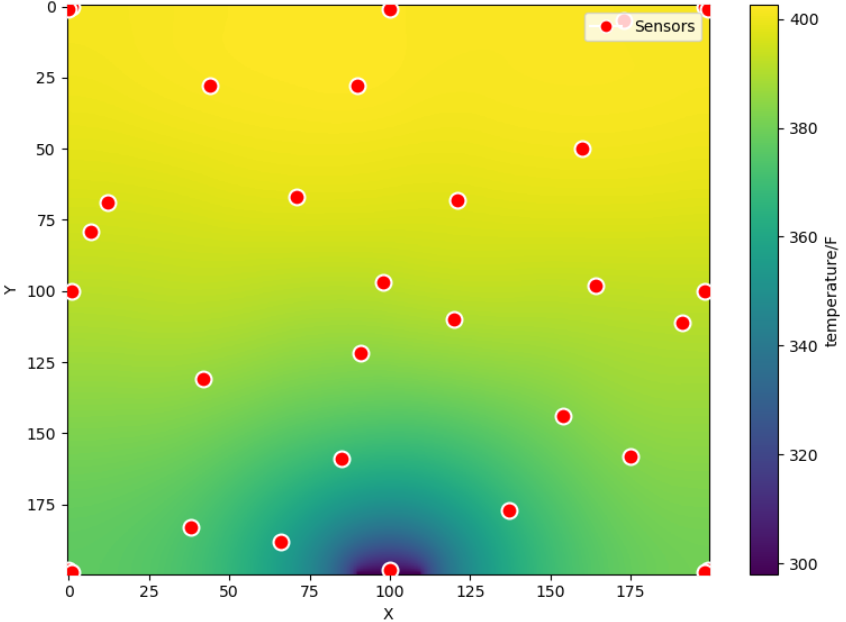}
		\caption{HSink condition}
		\label{fig:HSinksensors}
	\end{subfigure}
	\caption{Sparse sensor array configurations for temperature field reconstruction under different operating conditions.}
	\label{fig:sensor_arrays}
\end{figure}

\subsection{Experimental Setup}
All numerical experiments were conducted on a high-performance computing server running Ubuntu 20.04 LTS, equipped with an NVIDIA GeForce RTX 4090 GPU and 32 GB of system memory. The deep learning models were implemented using PyTorch 1.12.1 under Python 3.10, with CUDA 12.3 enabled for GPU acceleration.

To comprehensively evaluate the effectiveness of the proposed MTL-FNO framework, we compare it against several representative baseline models, including U-net\cite{Olaf2015U-Net}, DeepONet\cite{Lu2021DeepONet}, FNO\cite{Li2021FNO}, Senseiver\cite{Santos2023Senseiver}, PhySense\cite{Ma2025PhySense}, all baselines are trained independently for each individual task. The MTL-FNO backbone follows the standard FNO configuration, with $k=16$ truncated Fourier modes, a channel dimension of $C=32$, and $L=4$ layers. Both the lifting and projection fully connected layers have hidden dimensions of 128. For the CP tensor decomposition in task-specific fine-tuning, the ranks are set to $R=8$, all nonlinear activation functions employ the Gaussian Error Linear Unit (GeLU). MTL-FNO is trained using the Adam optimizer for up to 100 epochs. To balance rapid initial convergence with refined late-stage optimization, the learning rate is initialized at 0.001 and decayed by a factor of 0.5 every 20 epochs via StepLR scheduler. Consider the high cost of acquiring high-fidelity data, all experiments are conducted under a few-shot conditions, with 100 training and 20 test samples per task. Model performance is quantitatively assessed using a comprehensive set of metrics, including

\begin{itemize}
	\item Mean Squared Error(MSE):
	\[
	\text{MSE} = \frac{1}{N} \sum_{i=1}^{N} (\hat{y}_i - y_i)^2,
	\]
	
	\item Mean Absolute Error(MAE):
	\[
	\text{MAE} = \frac{1}{N} \sum_{i=1}^{N} |\hat{y}_i - y_i|,
	\]
	
	\item Coefficient of Determination($\mathbf{R}^2$):
	\[
	\textbf{R}^2 = 1 - \frac{\sum_{i=1}^{N} (\hat{y}_i - y_i)^2}{\sum_{i=1}^{N} (y_i - \bar{y})^2}.
	\]
\end{itemize}
In addition to accuracy metrics, per-sample inference time, total number of trainable parameters and floating-point operations (GFLOPs) are also reported to holistically evaluate computational efficiency and assess the potential for deployment on resource-constrained platforms.

\subsection{Experimental Results}

\subsubsection{Main Results of Case A}
\label{subsubsec:MR}
Table~\ref{tab:case_a_results} summarizes the comparative results of MTL-FNO against multiple baseline models in Case A. Several key observations can be drawn:
(i) \textbf{Model size}: Neural operator-based models, such as DeepONet and FNO, exhibit significantly fewer parameters than dense deep learning architectures including U-Net, Senseiver, and PhySense, owing to their streamlined network designs. MTL-FNO, leveraging a hard parameter sharing based multi-task framework, employs a unified model to handle all five tasks simultaneously, resulting in a total parameter count of 1.30 M. Compared with deploying five independent FNOs (5 $\times$ 1.09 M = 5.45 M), MTL-FNO reduces the total parameter count by approximately 76\%. This substantial compression is primarily attributed to the low-rank fine-tuning terms constructed via CP decomposition, which introduce only minimal task-specific parameters while maintaining expressive capacity.
(ii) \textbf{Computational overhead}: FNO achieves the lowest inference time and GFLOPs among all methods, primarily due to its use of FFT combined with low-frequency truncation, which drastically reduces computational complexity. MTL-FNO incurs a modest increase in inference latency (9.65 ms per sample) owing to the low-rank adaptations and slice-wise Cayley transformations, yet this remains well within the acceptable range for engineering applications. In contrast, Senseiver, built upon a Transformer architecture, requires approximately 10 ms per sample, while PhySense, which adopts a generative modeling paradigm, exceeds 100 ms per sample, rendering it impractical for real-time reconstruction.
(iii) \textbf{Prediction accuracy}: The baseline models exhibit notable performance deficiencies across different tasks. For instance, Senseiver struggles with pressure field P reconstruction ($\mathbf{R}^2$ = 0.689), PhySense severely underperforms on temperature field T prediction (negative $\mathbf{R}^2$), and vanilla FNO shows significant errors in normal stress field $\tau_{xx}$ reconstruction ($\mathbf{R}^2$ = -4.121). In contrast, MTL-FNO maintains consistently high accuracy across all five tasks, achieving the best or near-best metrics in every case, with particularly pronounced advantages in reconstructing stress components $\tau_{xx}$ and $\tau_{xy}$ ($\mathbf{R}^2$ of 0.936 and 0.993, respectively, versus FNO's -4.121 and 0.880). This strongly validates that the multi-task joint learning mechanism effectively captures cross-field commonalities from limited high-fidelity samples, thereby significantly enhancing generalization under few-shot conditions.

\begin{figure}[htbp]
	\centering

	\begin{subfigure}[b]{1.0\textwidth}
		\centering
		\includegraphics[width=\textwidth]{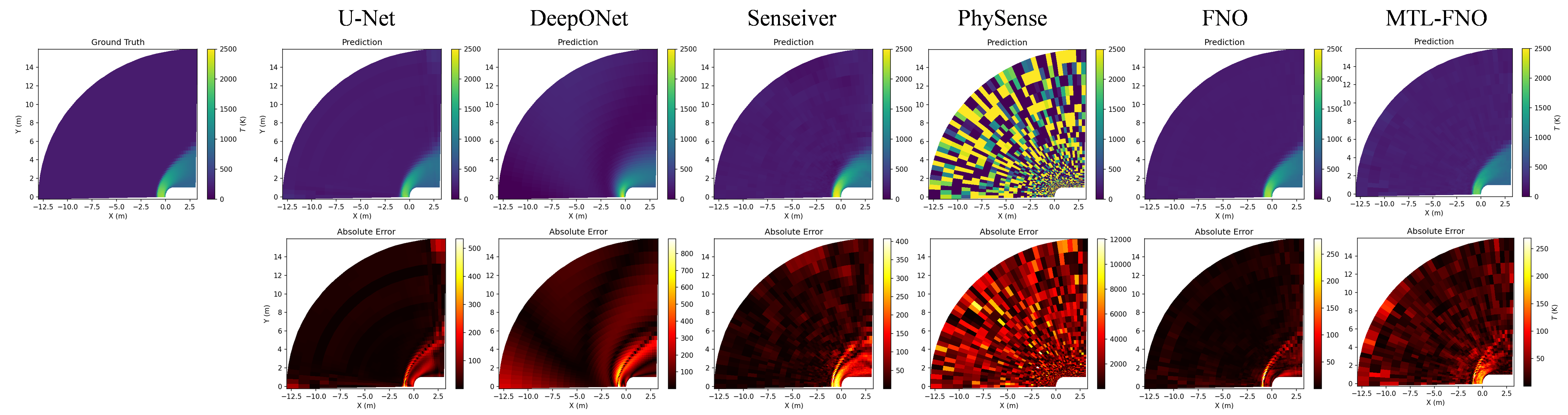}
		\caption{Temperature field T.}
		\label{fig:caseat}
	\end{subfigure}
	
	\begin{subfigure}[b]{1.0\textwidth}
		\centering
		\includegraphics[width=\textwidth]{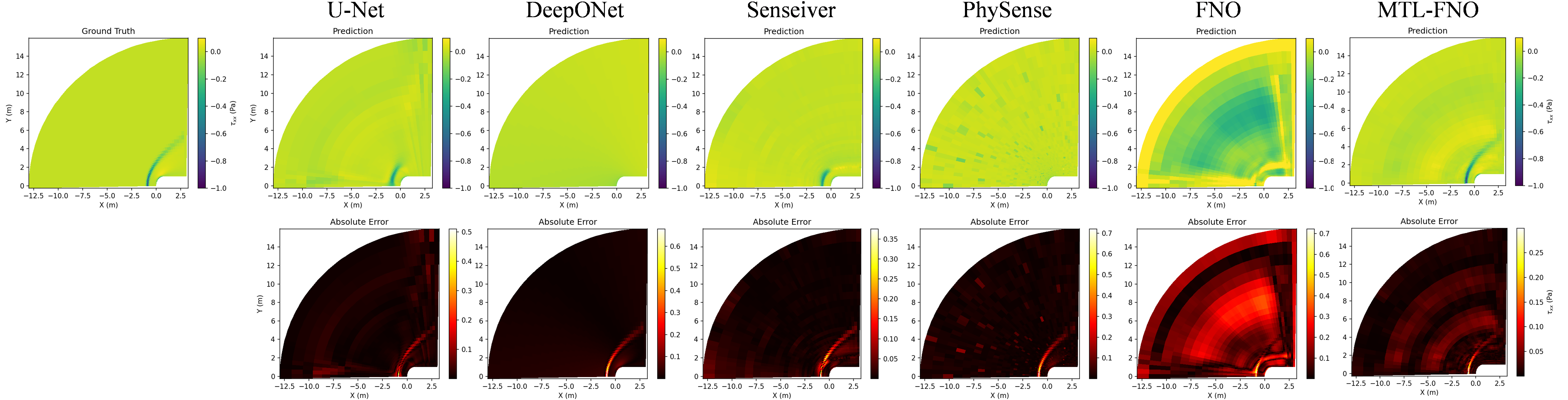}
		\caption{Stress field $\tau_{xx}$.}
		\label{fig:caseatxx}
	\end{subfigure}
	
	\begin{subfigure}[b]{1.0\textwidth}
		\centering
		\includegraphics[width=\textwidth]{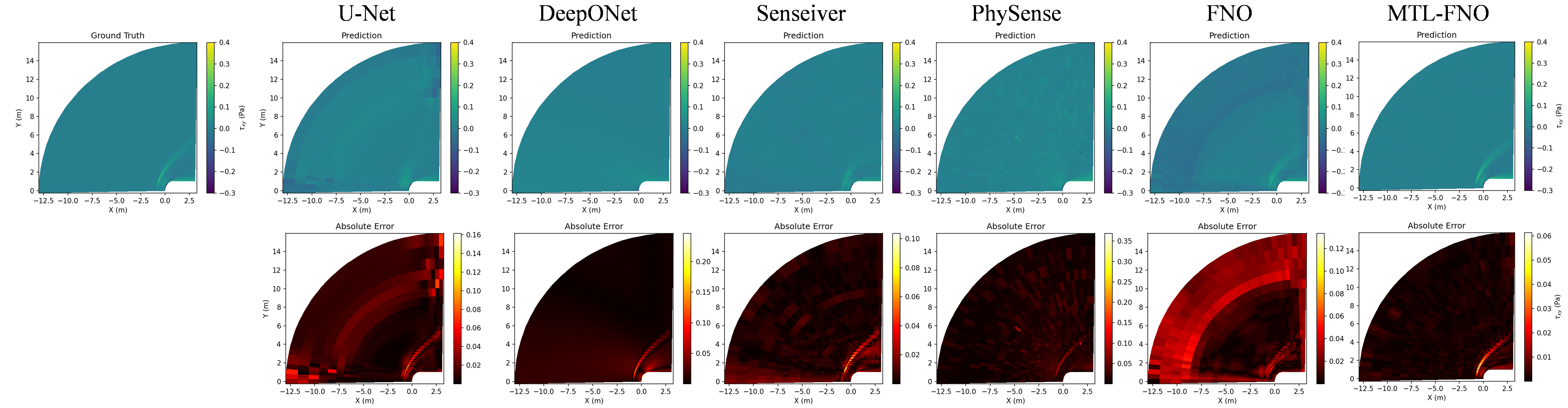}
		\caption{Stress field $\tau_{xy}$.}
		\label{fig:caseatxy}
	\end{subfigure}

	\caption{Ground truth of representative samples from Case A under different tasks, along with the reconstructed physical fields generated by various models and their corresponding error maps.}
	\label{fig:caseashow}
\end{figure}

Representative reconstructed fields and corresponding error maps for Case A are visualized in Fig.~\ref{fig:caseashow}. The predictions produced by MTL-FNO exhibit excellent agreement with the reference solutions, featuring minimal error magnitudes and no systematic bias. In comparison, other methods commonly suffer from local distortions or global offsets, further corroborating the superior accuracy and robustness of the proposed approach.

\begin{table}[htbp]
	\centering
	\caption{Comparisons between MTL-FNO and various baseline models in Case A.}
	\label{tab:case_a_results}
	\begin{tabular}{llccccccc}
		\toprule
		{Task} & {Model} & {Params (M)} & {GFLOPs} & {\makecell{Inference \\ Time \\(ms/sample)}} & {MSE($\downarrow$)} & {MAE($\downarrow$)} & {$\mathbf{R}^2$($\uparrow$)} \\
		\midrule
		\multirow{6}{*}{P}
		 & U-net    & 13.39 & 3.452 & 13.23 & 0.040 & 0.111 & 0.991 \\
		& DeepONet & $\bm{0.51}$   & 2.00     & $\bm{0.95}$ & 0.511 & 0.467 & 0.887 \\
		& Senseiver & 11.42 & 2.21 & 10.20 & 1.404 & 0.667 & 0.689 \\
		& PhySense      & 5.59  & 12.78 & 108.54 & 0.060 & 0.135 & 0.975 \\
		& FNO      & 1.09  & $\bm{0.19}$ & 1.57 & 0.012 & 0.066 & 0.997 \\
		& MTL-FNO  & 1.30 & 0.41 & 9.65 & $\bm{0.010}$ & $\bm{0.059}$ & $\bm{0.998}$ \\
		\midrule
		\multirow{6}{*}{T}
		& U-net    & 13.39 & 3.45 & 15.290 & 19167.430 & 87.101 & 0.947 \\
		& DeepONet & 0.51   & 2.00     & 5.146 & 42399.988 & 139.268 & 0.884 \\
		& Senseiver & 11.42 & 2.21 & 13.359 & 38481.008 & 140.264 & 0.895 \\
		& PhySense      & 5.59  & 12.78 & 118.12 & 8947156.700 & 2378.049 & -30.291 \\
		& FNO      & 1.09  & 0.19 & $\bm{1.342}$ & 22277.975 & 96.446 & 0.939 \\
		& MTL-FNO  & / & 0.41 & 9.65 & $\bm{10225.969}$ & $\bm{66.875}$ & $\bm{0.972}$ \\
		\midrule
		\multirow{6}{*}{$\tau_{xx}$}
		& U-net    & 13.39 & 3.45 & 16.02 & 0.0020 & 0.0242 & 0.715 \\
		& DeepONet & 0.51   & 2.00     & 5.10 & 0.0057 & 0.0401 & 0.186 \\
		& Senseiver & 11.42 & 2.21 & 10.28 & 0.0022 & 0.0245 & 0.682 \\
		& PhySense      & 5.59  & 12.78 & 120.00 & 0.0068 & 0.0422 & -0.082 \\
		& FNO      & 1.09  & 0.19 & $\bm{0.42}$ & 0.0359 & 0.1513 & -4.121 \\
		& MTL-FNO  & / & 0.41 & 9.65 & $\bm{0.0005}$ & $\bm{0.0159}$ & $\bm{0.936}$ \\
		\midrule
		\multirow{6}{*}{$\tau_{xy}$}
		& U-net    & 13.39 & 3.45 & 16.52 & 0.00081 & 0.0188 & 0.853 \\
		& DeepONet & 0.51   & 2.00     & 4.06 & 0.00282 & 0.0346 & 0.511 \\
		& Senseiver & 11.42 & 2.21 & 14.05 & 0.00074 & 0.0167 & 0.874 \\
		& PhySense      & 5.59  & 12.78 & 122.00 & 0.00105 & 0.0214 & 0.741 \\
		& FNO      & 1.09  & 0.19 & $\bm{0.98}$ & 0.00075 &0.0192 & 0.880 \\
		& MTL-FNO  & / & 0.41 & 9.65 & $\bm{0.00004}$ & $\bm{0.0040}$ & $\bm{0.993}$ \\
		\midrule
		\multirow{6}{*}{$\tau_{yy}$} 
		& U-net    & 13.39 & 3.45 & 18.34 & 0.00030 & 0.0116 & 0.968 \\
		& DeepONet & 0.51   & 2.00     & 5.51 & 0.00345 & 0.0397 & 0.631 \\
		& Senseiver & 11.42 & 2.21 & 17.43 & 0.00139 & 0.0238 & 0.852 \\
		& PhySense      & 5.59  & 12.78 & 121.94 & 0.00037 & 0.0132 & 0.953 \\
		& FNO      & 1.09  & 0.19 & $\bm{1.61}$ & 0.00008 & 0.0061 & 0.991 \\
		& MTL-FNO  & / & 0.41 & 9.65 & $\bm{0.00007}$ & $\bm{0.0054}$ & $\bm{0.993}$ \\
		\bottomrule
	\end{tabular}
\end{table}

\subsubsection{Main Results of Case B}
Table~\ref{tab:case_b_results} reports the comparative results in Case B. The key findings are summarized as follows:
(i) \textbf{Model size}: Consistent with Case A, MTL-FNO demonstrates significant parameter efficiency in Case B. By sharing the backbone across three thermal scenarios, MTL-FNO requires only 1.21 M parameters, compared with 3.27 M for three independent FNOs—a reduction of approximately 60\%. Notably, despite the increase in task diversity from Case A to Case B, the parameter count of MTL-FNO grows only marginally (from 1.30 M to 1.21 M, with the slight decrease reflecting the smaller number of tasks). This further confirms that the low-rank fine-tuning terms effectively decouple model capacity from the number of tasks, making MTL-FNO highly scalable for multi-field reconstruction.
(ii) \textbf{Computational overhead}: The trend in computational efficiency mirrors that of Case A. FNO maintains the lowest inference time and GFLOPs, while MTL-FNO incurs a modest overhead (6.08 ms per sample, 1.93 GFLOPs) due to the Cayley transform and low-rank adaptations. Senseiver exhibits substantially higher latency (up to 38.01 ms) owing to the denser grid resolution. PhySense again proves unsuitable for real-time applications, with inference times exceeding 1100 ms per sample.
(iii) \textbf{Prediction accuracy}: MTL-FNO demonstrates remarkable robustness and balance across the three diverse thermal scenarios. Under the ADlet condition, it comprehensively outperforms all baselines, achieving the lowest MSE (0.371) and the highest $\mathbf{R}^2$ (0.970). For the DSine scenario, MTL-FNO achieves $\mathbf{R}^2$ of 0.988, marginally surpassing FNO (0.986), while its MSE (0.493) remains extremely close to FNO's best value (0.491). Under the HSink condition, MTL-FNO attains $\mathbf{R}^2$ of 0.986, on par with U-Net (0.986), with only a slight increase in MSE relative to the best performer. This balanced performance across heterogeneous thermal boundary conditions underscores the effectiveness of the proposed multi-task framework in extracting shared physical representations while preserving task-specific features.

\begin{figure}[htbp]
	\centering
	
	\begin{subfigure}[b]{1.0\textwidth}
		\centering
		\includegraphics[width=\textwidth]{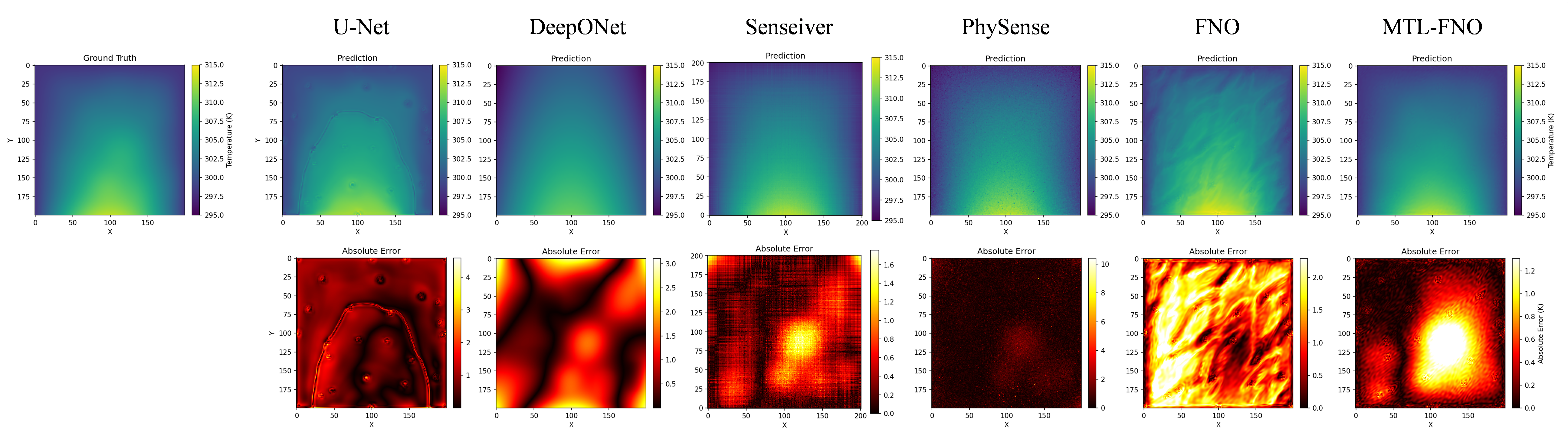}
		\caption{ADlet.}
		\label{fig:casebadlt}
	\end{subfigure}
	
	\begin{subfigure}[b]{1.0\textwidth}
		\centering
		\includegraphics[width=\textwidth]{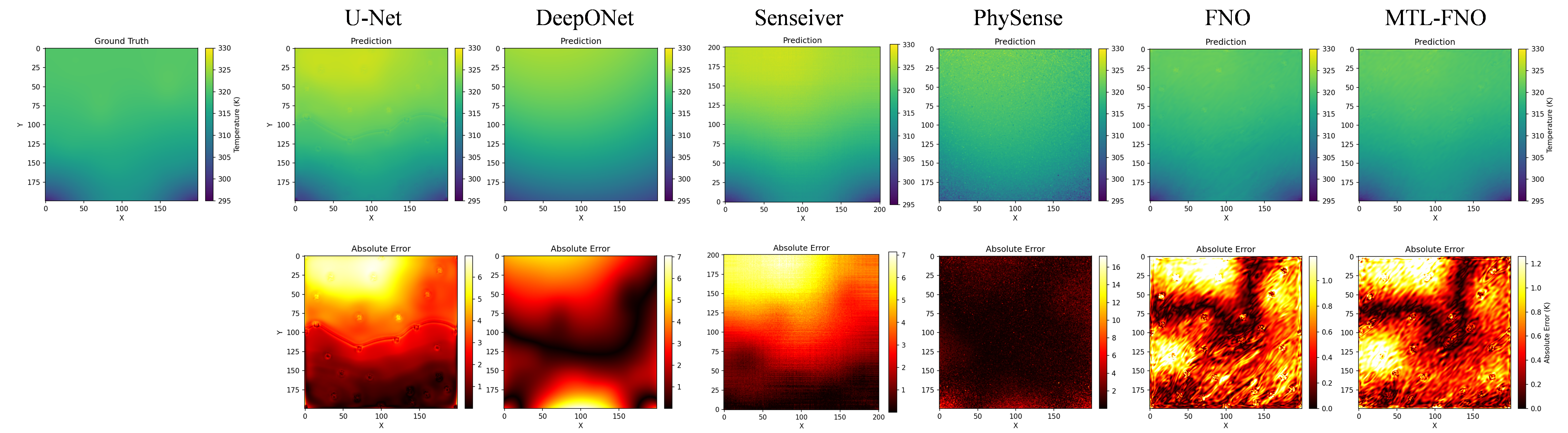}
		\caption{DSine.}
		\label{fig:casebdsine}
	\end{subfigure}
	
	\begin{subfigure}[b]{1.0\textwidth}
		\centering
		\includegraphics[width=\textwidth]{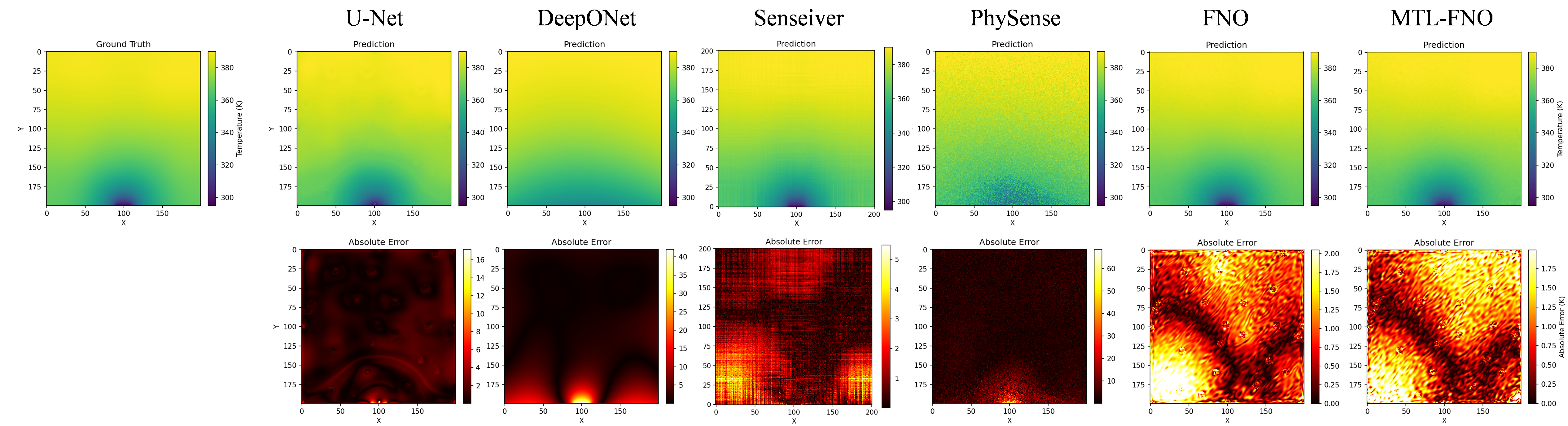}
		\caption{HSink.}
		\label{fig:casebhsink}
	\end{subfigure}
	
	\caption{Ground truth of representative samples from Case B under different tasks, along with the reconstructed physical fields generated by various models and their corresponding error maps.}
	\label{fig:casebshow}
\end{figure}

Representative reconstructed fields and error maps for Case B are shown in Fig.~\ref{fig:casebshow}. MTL-FNO consistently produces reconstructions that closely match the ground truth, with error maps exhibiting negligible magnitudes and no discernible structured artifacts. In contrast, baselines display varying degrees of local distortion, particularly in regions with sharp thermal gradients, further validating the accuracy and generalization capability of MTL-FNO.

\begin{table}[htbp]
	\centering
	\caption{Case B: comparisons between MTL-FNO and various baseline models.}
	\label{tab:case_b_results}
	\begin{tabular}{llccccccc}
		\toprule
		{Task} & {Model} & {Params (M)} & {GFLOPs} & {\makecell{Inference \\ Time \\(ms/sample)}} & {MSE($\downarrow$)} & {MAE($\downarrow$)} & {\textbf{R\textsuperscript{2}}($\uparrow$)} \\
		\midrule
		\multirow{6}{*}{ADlet} 
		& U-Net & 13.39 & 37.71 &  3.03 & 1.014 & 0.839 & 0.914 \\
		& DeepONet & $\bm{0.51}$ & 10.37 & 11.00 & 1.134 & 0.862 & 0.906 \\
		& Senseiver & 11.43 & 17.68 & 23.36 & 0.405 & 0.487 & 0.968 \\
		& PhySense & 5.59 & 269.58 & 1122.79 & 0.679 & 0.638 & 0.944 \\
		& FNO & 1.09 & $\bm{1.87}$ & $\bm{1.58}$ & 2.14 & 1.319 & 0.825 \\
		& MTL-FNO & 1.21 & 1.93  & 6.08 & $\bm{0.371}$ & $\bm{0.439}$ & $\bm{0.970}$ \\
		\midrule
		\multirow{6}{*}{DSine} 
		& U-Net & 13.39 & 37.71 & 2.50  & 6.268 & 1.794 &  0.798 \\
		& DeepONet & 0.51 & 10.37 & 5.79 & 3.830 & 1.477 & 0.806 \\
		& Senseiver & 11.43 & 17.68 & 38.01 & 6.374 & 1.801 & 0.843 \\
		& PhySense & 5.59 & 269.58 & 1128.15 & 3.125 & 0.487 & 0.968 \\
		& FNO & 1.09 & 1.87 & $\bm{1.47}$ & $\bm{0.491}$ & $\bm{0.545}$ & 0.986 \\
		& MTL-FNO & / & 1.93  & 6.08 & 0.493 & 0.550 & $\bm{0.988}$ \\
		\midrule
		\multirow{6}{*}{HSink} 
		& U-Net & 13.39 & 37.71 & 9.69 & $\bm{3.811}$ & $\bm{1.508}$ & $\bm{0.986}$ \\
		& DeepONet & 0.51 & 10.37 & $\bm{5.84}$ & 17.845 & 2.630 & 0.847 \\
		& Senseiver & 11.43 & 17.68 & 37.53 & 11.580 & 2.597 & 0.953 \\
		& PhySense & 5.59 & 269.58 & 1131.00 & 25.466 & 3.432 & 0.771 \\
		& FNO & 1.09 & 1.87 & 7.70 & 4.624 & 1.702 & 0.963 \\
		& MTL-FNO & / & 1.93  & 6.08 & 4.448 & 1.661 & 0.986\\
		\bottomrule
	\end{tabular}
\end{table}

\subsubsection{Further Discussion}
\label{subsubsec:ablation}
\textbf{Ablation study:} To systematically investigate the impact of layer-wise parameter sharing, polar decomposition and the Cayley transform on the performance of MTL-FNO, we constructs three ablation variants evaluated on both Case A and B:
\begin{itemize}
	\item \textbf{MTL-FNO-noshare} removes the layer-wise shared and task-specific parameter partitioning of MTL-FNO, i.e., the model architecture degenerates to a standard FNO. Unlike independently training single-field FNO in \cref{subsubsec:MR}, MTL-FNO-noshare is trained on data from all tasks as a unified framework.
	\item \textbf{MTL-FNO-nopolar} omits polar decomposition and directly applies task-specific lightweight adaptation $\Delta \mathcal{R}_i^l$, constructed via CP decomposition, to the shared $\mathcal{R}_{\text{share}}^l$.
	\item \textbf{MTL-FNO-nocayley} omits the Cayley transform. After adding the task-specific adaptation $\Delta \mathcal{K}_i^l$ to the shared $\mathcal{K}_{\text{share}}^l$, it directly perform slice-wise multiplication with the positive semi-definite tensor.
\end{itemize}

\begin{figure}[htbp]
	\centering
	\begin{subfigure}[b]{0.3\linewidth}
		\centering
		\includegraphics[width=\linewidth]{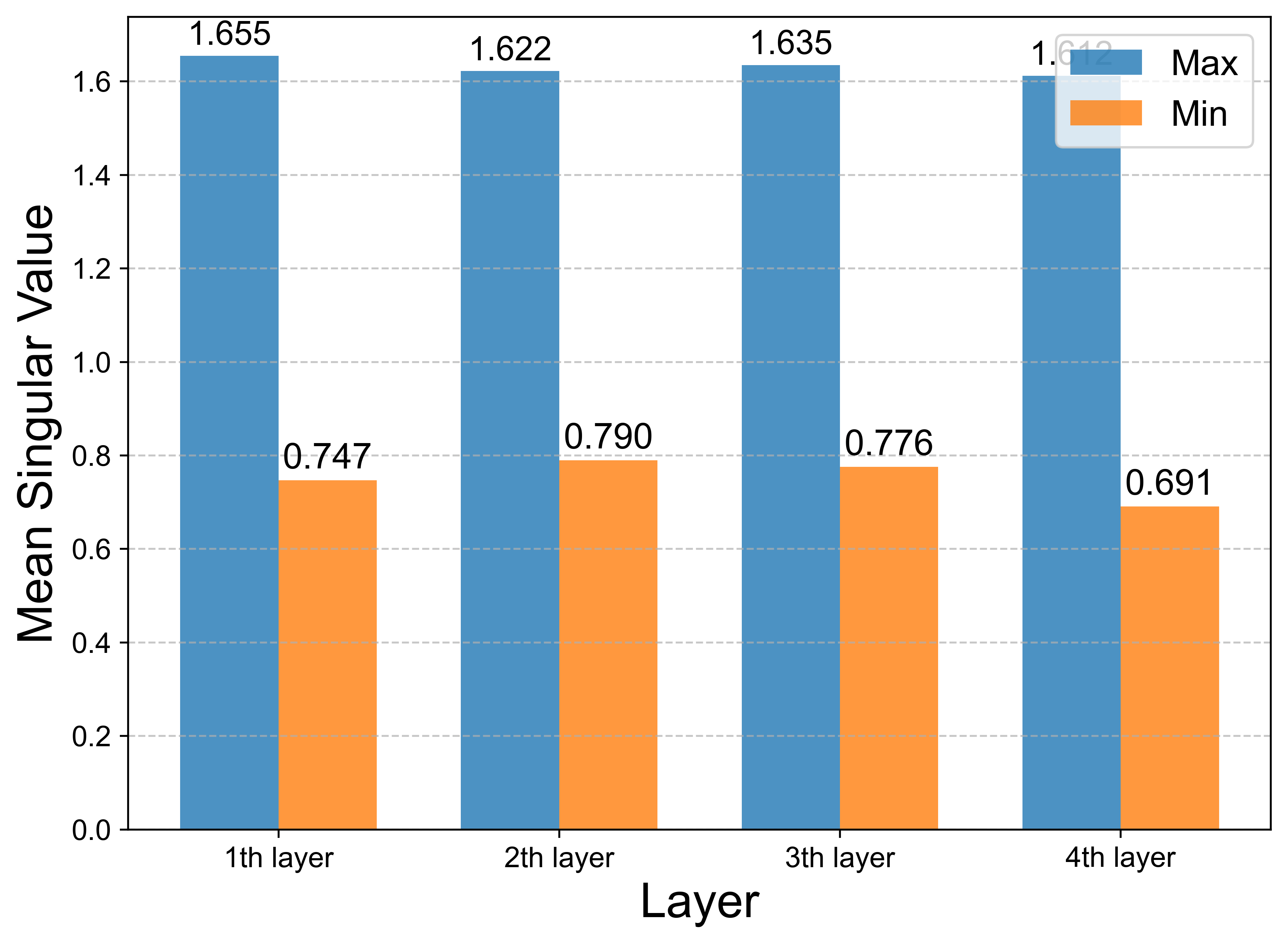}
		\caption{Case A}
		\label{fig:case_a_singularvalue}
	\end{subfigure}
	\hspace{3em}%
	\begin{subfigure}[b]{0.3\linewidth}
		\centering
		\includegraphics[width=\linewidth]{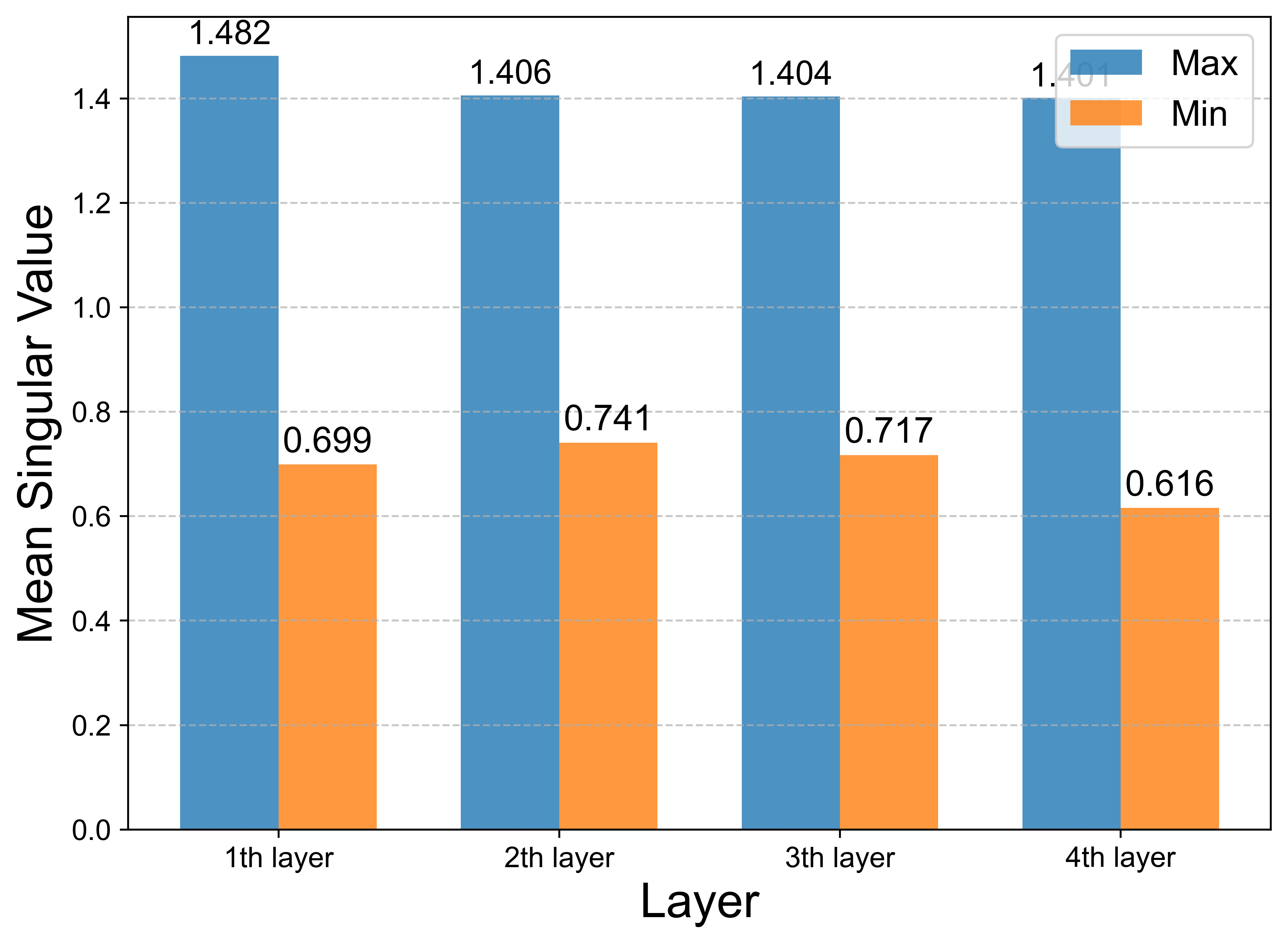}
		\caption{Case B}
		\label{fig:case_b_singularvalue}
	\end{subfigure}
	\caption{The average maximum and minimum singular value statistics of 'unitary' matrix slices in MTL-FNO-nocayley for Case A and Case B.}
	\label{fig:singularvalue_comparison}
\end{figure}

The results are summarized in Table~\ref{tab:case_a_results_polarcayley} and~\ref{tab:case_b_results_polar_cayley}. All four models have comparable parameter budgets and inference costs. MTL‑FNO‑noshare is the lightest (1.09 M parameters, 0.19 GFLOPs in Case A, 0.976 ms inference in Case B), while MTL‑FNO and MTL‑FNO‑nocayley carry a slight overhead due to the additional positive semi‑definite tensors. In reconstruction accuracy, MTL‑FNO consistently achieves the best results across all tasks, often by a substantial margin. (i) Comparison with MTL‑FNO‑noshare evaluates the effect of layer‑wise shared and task‑specific partitioning. On the relatively easier tasks (e.g., pressure field P in Case A, DSine and HSink in Case B), the variant performs reasonably well. However, when task conflicts are severe, the performance drops markedly: for stress field $\tau_{xx}$ in Case A, MTL‑FNO reaches $\textbf{R}^2=0.936$ whereas MTL‑FNO‑noshare collapses to $\textbf{R}^2=-0.730$; similarly, on stress field $\tau_{xy}$ the $\textbf{R}^2$ falls from 0.993 to 0.631 and MSE increases more than tenfold. This stark contrast confirms that layer‑wise separation of shared and task‑specific parameters is crucial for capturing common physical features across fields while preserving task‑specific characteristics, thereby enabling MTL‑FNO to maintain consistently high accuracy on all tasks. (ii) Comparison with MTL‑FNO‑nopolar highlights the role of polar decomposition. Without polar decomposition, the model cannot decouple amplitude scaling from phase shifting, leading to unresolved task interference. Consequently, its accuracy becomes highly imbalanced across tasks. For instance, on stress field $\tau_{xy}$ in Case A, $\textbf{R}^2$ drops from 0.993 to 0.576, and on ADlet in Case B, it falls from 0.970 to 0.904. (iii) Comparison with MTL‑FNO‑nocayley underscores the necessity of the Cayley transform. Omitting the Cayley transform breaks the decoupled optimization of phase and amplitude. MTL‑FNO‑nocayley omits the Cayley transform, thereby breaking the decoupled optimization of phase and amplitude. As shown in Fig.~\ref{fig:singularvalue_comparison}, the mean maximum and minimum singular value statistics of the 'unitary' matrices in MTL-FNO-nocayley deviate significantly from unity——the ideal value for a perfectly unitary matrix. This loss of unitary geometric fidelity destroys the decoupled optimization of phase and amplitude, leading to severe degradation. In Case A, $\textbf{R}^2$ for temperature field T collapses to 0.414, and for field $\tau_{xx}$ it becomes negative (–1.458). Even in Case B, performance on ADlet and HSink degrades noticeably ($\textbf{R}^2$ of 0.938 and 0.953, compared to 0.970 and 0.986 for MTL‑FNO). 

In summary, each component—layer‑wise sharing, polar decomposition, and the Cayley transform—plays a necessary role in the performance of MTL‑FNO. Their combined effect yields robust multi‑task learning with high accuracy and balanced performance across all physical fields. 

\begin{figure}[htbp]
	\centering
	\begin{subfigure}[b]{0.3\linewidth}
		\centering
		\includegraphics[width=\linewidth]{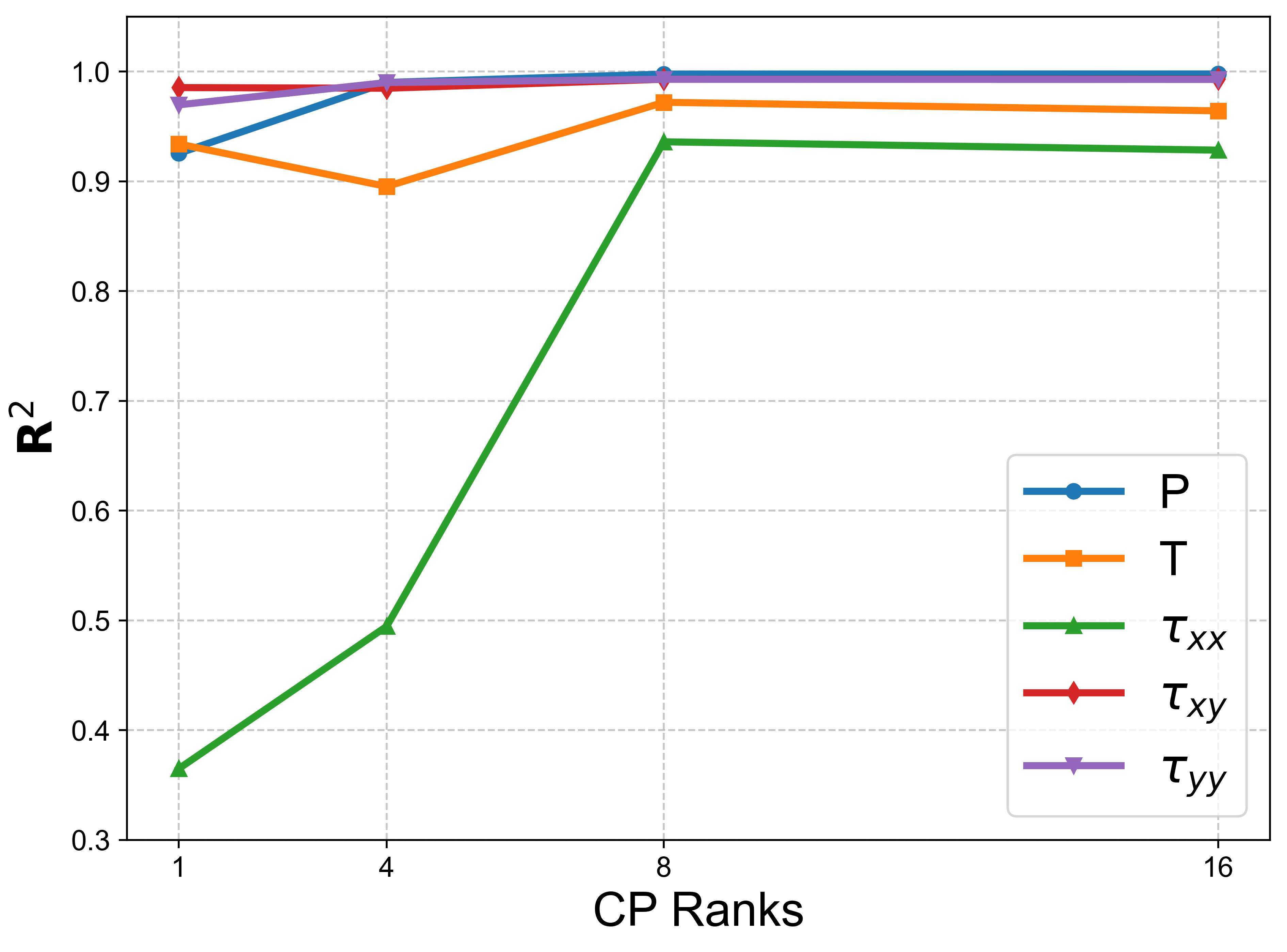}
		\caption{Case A}
		\label{fig:case_a_rank}
	\end{subfigure}
	\hspace{3em}%
	\begin{subfigure}[b]{0.3\linewidth}
		\centering
		\includegraphics[width=\linewidth]{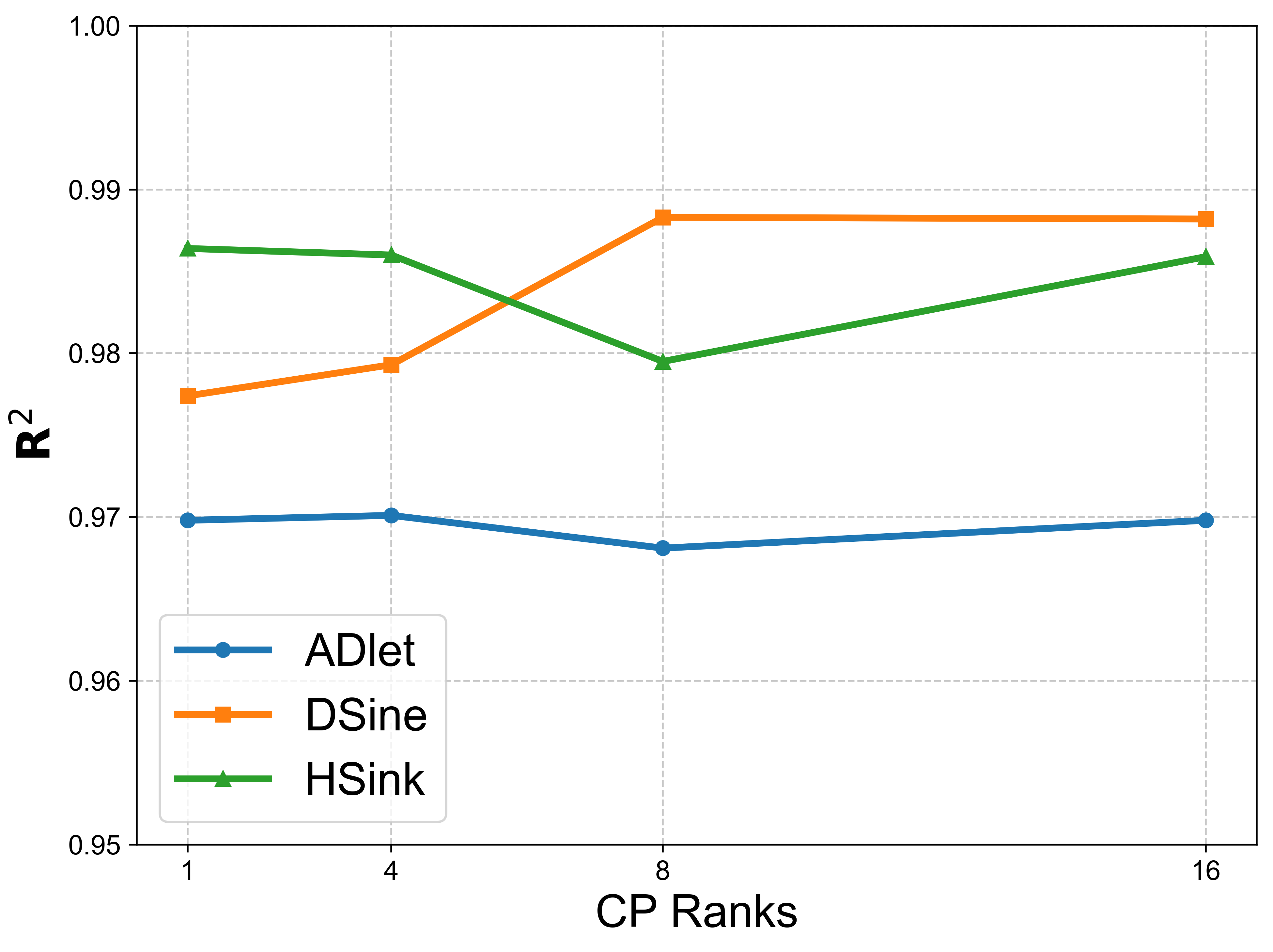}
		\caption{Case B}
		\label{fig:case_b_rank}
	\end{subfigure}
	\caption{$\mathbf{R}^2$ Performance of MTL-FNO with Different CP Ranks for Case A and Case B.}
	\label{fig:combined_rank}
\end{figure}

\textbf{The effects of CP ranks:} This paper investigates the impact of CP ranks on MTL-FNO by setting $R$ to 1, 4, 8, and 16. The experimental results are presented in Fig.~\ref{fig:combined_rank}. In Case A, it is evident that the model achieves its worst performance when $R=1$, and improves its performance significantly as $R$ increases. For instance, in stress field $\tau_{xx}$, $\textbf{R}^2$ rises from 0.365 to 0.936 when $R$ grows from 1 to 8. When $R \geq 8$, the performance of MTL-FNO saturates and remains at a consistently high level. In contrast, for Case B, MTL-FNO already delivers strong accuracy at $R=1$ (with $\textbf{R}^2= 0.970$ in ADlet, $\textbf{R}^2=0.977$ in DSine, and $\textbf{R}^2=0.986$ in HSink, respectively), and continues to maintain excellent performance as $R$ increases, with only minor fluctuations. This contrast suggests that a small CP rank may not adequately preserve or exploit task-specific features for some tasks, while it proves sufficient for others. Considering both model performance and parameter efficiency, this paper select $R=8$ as the default configuration. 

\begin{table}[htbp]
	\centering
	\caption{Case A: the effects of layer-wise parameter sharing, polar decomposition and the Cayley transform on MTL-FNO}
	\label{tab:case_a_results_polarcayley}
	\begin{tabular}{llccccccc}
		\toprule
		{Task} & {Model} & {Params (M)} & {GFLOPs} & {\makecell{Inference \\ Time \\(ms/sample)}} & {MSE($\downarrow$)} & {MAE($\downarrow$)} & {\textbf{R\textsuperscript{2}}($\uparrow$)} \\
		\midrule
		\multirow{4}{*}{P}
		& MTL-FNO           & 1.30 & 0.41 & 9.65 & $\bm{0.010}$ & $\bm{0.059}$ & $\bm{0.998}$ \\
		& MTL-FNO-noshare   & $\bm{1.09}$ & $\bm{0.19}$ & $\bm{1.60}$ & 0.037 & 0.141 & 0.992 \\
		& MTL-FNO-nopolar   & 1.26 & 0.31 & 3.17 & 0.246 & 0.456 & 0.946 \\
		& MTL-FNO-nocayley  & 1.30 & 0.32 & 5.24 & 0.159 & 0.377 & 0.942 \\
		\midrule
		\multirow{4}{*}{T}
		& MTL-FNO           & / & 0.41 & 9.65 & $\bm{10225.969}$ & $\bm{66.875}$ & $\bm{0.972}$ \\
		& MTL-FNO-noshare   & / & $\bm{0.19}$ & $\bm{1.60}$ & 21944.293 & 122.036 & 0.940 \\
		& MTL-FNO-nopolar   & / & 0.31 & 3.17 & 38417.348 & 140.020 & 0.895 \\
		& MTL-FNO-nocayley  & / & 0.32 & 5.24 & 164913.984 & 350.304 & 0.414 \\
		\midrule
		\multirow{4}{*}{$\tau_{xx}$}
		& MTL-FNO           & / & 0.41 & 9.65 & $\bm{0.0005}$ & $\bm{0.0159}$ & $\bm{0.936}$ \\
		& MTL-FNO-noshare   & / & $\bm{0.19}$ & $\bm{1.60}$ & 0.0121 & 0.1050 & -0.730 \\
		& MTL-FNO-nopolar   & / & 0.31 & 3.17 & 0.0059 & 0.0678 & 0.157 \\
		& MTL-FNO-nocayley  & / & 0.32 & 5.24 & 0.0113 & 0.0868 & -1.458 \\
		\midrule
		\multirow{4}{*}{$\tau_{xy}$}
		& MTL-FNO           & / & 0.41 & 9.65 & $\bm{0.00004}$ & $\bm{0.0040}$ & $\bm{0.993}$ \\
		& MTL-FNO-noshare   & / & $\bm{0.19}$ & $\bm{1.60}$ & 0.0021 & 0.0440 & 0.631 \\
		& MTL-FNO-nopolar   & / & 0.31 & 3.17 & 0.00240 & 0.0449 & 0.576 \\
		& MTL-FNO-nocayley  & / & 0.32 & 5.24 & 0.00192 & 0.0410 & 0.485 \\
		\midrule
		\multirow{4}{*}{$\tau_{yy}$}
		& MTL-FNO           & / & 0.41 & 9.65 & $\bm{0.00007}$ & $\bm{0.0054}$ & $\bm{0.993}$ \\
		& MTL-FNO-noshare   & / & $\bm{0.19}$ & $\bm{1.60}$ & 0.00013 & 0.0091 & 0.986 \\
		& MTL-FNO-nopolar   & / & 0.31 & 3.17 & 0.00015 & 0.0107 & 0.984 \\
		& MTL-FNO-nocayley  & / & 0.32 & 5.24 & 0.00181 & 0.0403 & 0.708 \\
		\bottomrule
	\end{tabular}
\end{table}

\begin{table}[htbp]
	\centering
	\caption{Case B: the effects of layer-wise parameter sharing, polar decomposition and the Cayley transform on MTL-FNO}
	\label{tab:case_b_results_polar_cayley}
	\begin{tabular}{llcccccc}
		\toprule
		{Task} & {Model} & {Params (M)} & {GFLOPs} & {\makecell{Inference \\ Time \\(ms/sample)}} & {MSE($\downarrow$)} & {MAE($\downarrow$)} & {\textbf{R\textsuperscript{2}}($\uparrow$)} \\
		\midrule
		\multirow{4}{*}{ADlet} 
		& MTL-FNO            & 1.21 & 1.93 & 6.083 & $\bm{0.371}$ & $\bm{0.439}$ & $\bm{0.970}$ \\
		& MTL-FNO-noshare    & $\bm{1.09}$ & $\bm{1.87}$   & $\bm{0.976}$ & 1.293 & 0.964 & 0.895 \\
		& MTL-FNO-nopolar    & 1.17 & 1.89 & 3.466 & 1.162 & 0.914 & 0.904 \\
		& MTL-FNO-nocayley   & 1.21 & 1.90 & 4.642 & 0.756 & 0.662 & 0.938 \\
		\midrule
		\multirow{4}{*}{DSine} 
		& MTL-FNO            & / & 1.93 & 6.083 & $\bm{0.493}$ & $\bm{0.550}$ & $\bm{0.988}$ \\
		& MTL-FNO-noshare    & / & $\bm{1.87}$   & $\bm{0.976}$   & 1.430 & 0.960 & 0.965 \\
		& MTL-FNO-nopolar    & / & 1.89 & 3.466 & 0.909 & 0.736 & 0.976 \\
		& MTL-FNO-nocayley   & / & 1.90 & 4.642 & 0.635 & 0.623 & 0.983 \\
		\midrule
		\multirow{4}{*}{HSink} 
		& MTL-FNO            & / & 1.93 & 6.083 & $\bm{4.448}$ & $\bm{1.661}$ & $\bm{0.986}$ \\
		& MTL-FNO-noshare    & / & $\bm{1.87}$   & $\bm{0.976}$   & 4.966 & 1.762 & 0.985 \\
		& MTL-FNO-nopolar    & / & 1.89 & 3.466 & 6.002 & 1.885 & 0.977 \\
		& MTL-FNO-nocayley   & / & 1.90 & 4.642 & 11.983 & 2.908 & 0.953 \\
		\bottomrule
	\end{tabular}
\end{table}

\begin{figure}[htbp]
	\centering
	\begin{subfigure}[b]{0.85\linewidth}
		\centering
		\includegraphics[width=\linewidth]{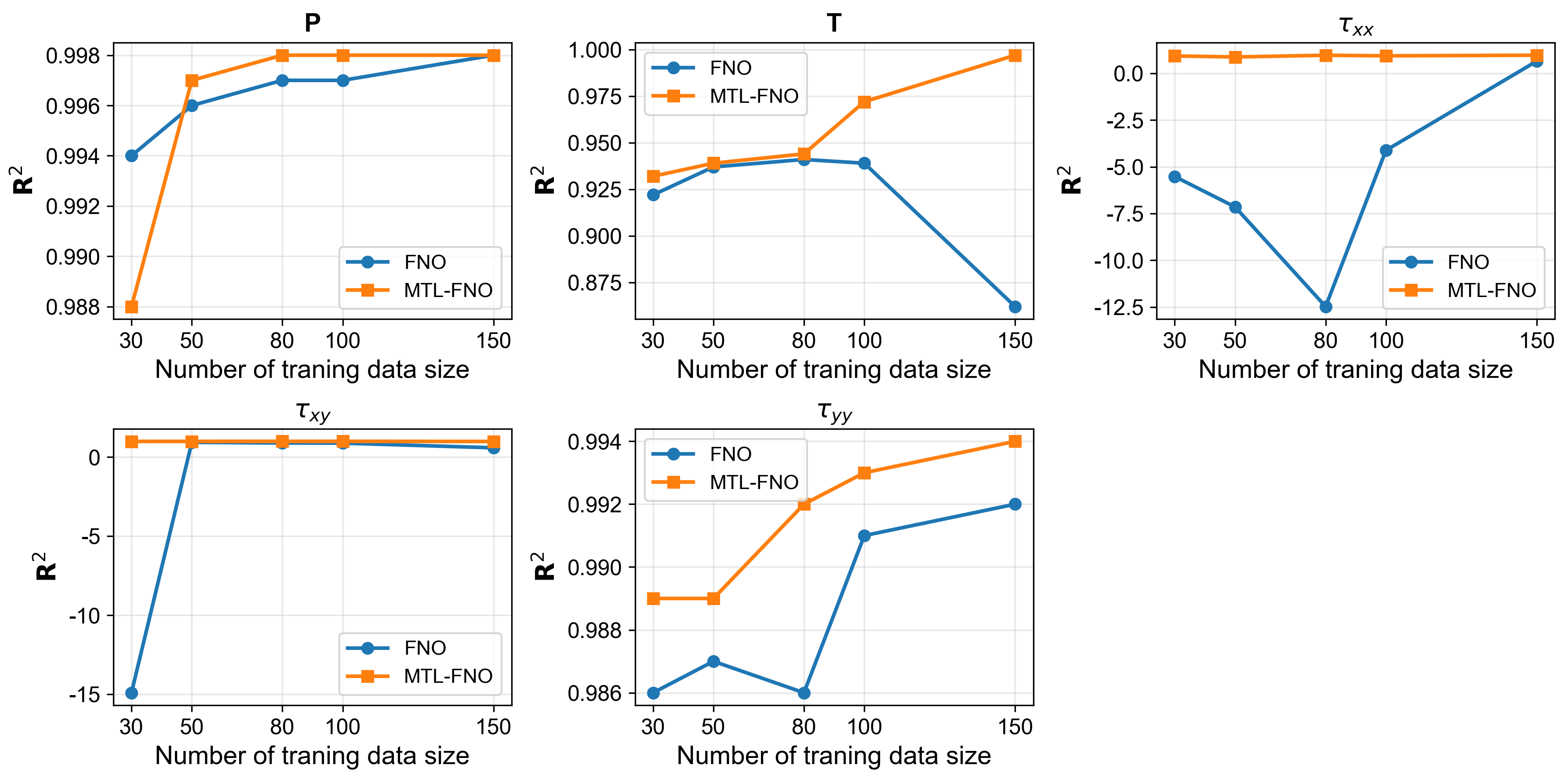}
		\caption{Case A}
		\label{fig:case_a_datasize}
	\end{subfigure}
	\hspace{3em}%
	\begin{subfigure}[b]{0.85\linewidth}
		\centering
		\includegraphics[width=\linewidth]{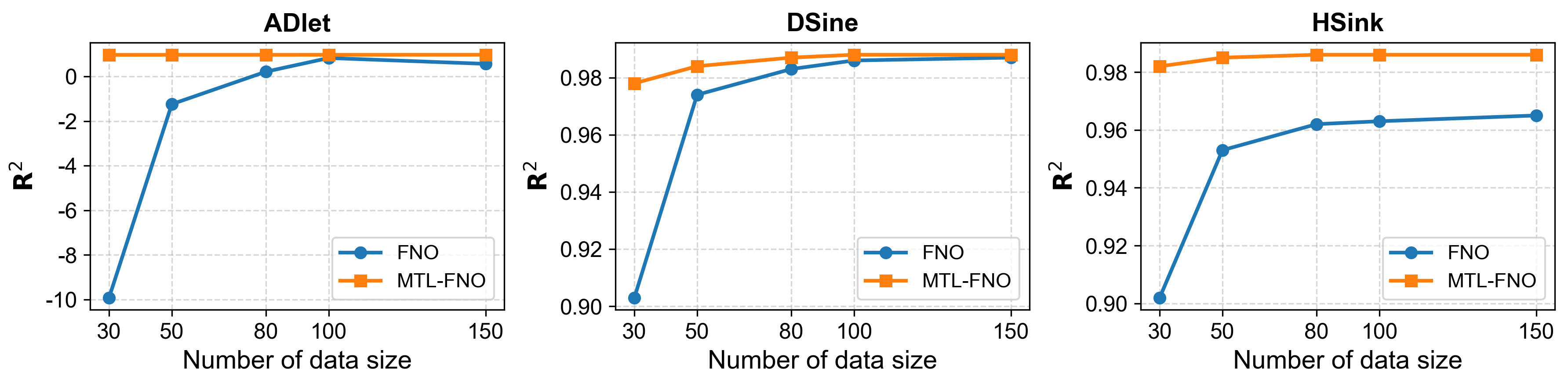}
		\caption{Case B}
		\label{fig:case_b_datasize}
	\end{subfigure}
	\caption{$\mathbf{R}^2$ Performance of FNO and MTL-FNO with Different Training Data Size for Case A and Case B.}
	\label{fig:combined_datasize}
\end{figure}

\textbf{The effects of training data size:} To thoroughly investigate the robustness of MTL-FNO under few-shot conditions, we trained it with varying numbers of training samples—30, 50, 80, 100, and 150—and compared its performance against the standard FNO on Case A and Case B, with the results summarized in Fig.\ref{fig:combined_datasize}. The resulting $\mathbf{R}^2$ reveal a pronounced advantage of multi-task learning under few-shot conditions. In Case A, FNO yields extremely negative $\mathbf{R}^2$ for the stress components $\tau_{xx}$ and $\tau_{xy}$ when trained on up to 100 samples (e.g., 
-14.906 for $\tau_{xy}$ at 30 samples), whereas MTL-FNO immediately lifts these metrics to 0.928 and 0.988, respectively, and further improves them to over 0.96 and 0.99 as more data become available. A similar pattern emerges in Case B, FNO completely fails on the ADlet scenarios with only 30 samples ($\mathbf{R}^2$ = -9.912) and remains unsatisfactory at 50 samples ($\mathbf{R}^2$ = -1.243), while MTL‑FNO achieves $\mathbf{R}^2$ of 0.967 with just 30 samples and maintains stable values around 0.97 across all sample sizes. Notably, FNO exhibits non‑monotonic and even deteriorating performance on several fields as the training set grows. For instance, the $\mathbf{R}^2$ of FNO on the temperature field T in Case A decreases from 0.941 (80 samples) to 0.862 (150 samples), and its score on $\tau_{xx}$ fluctuates dramatically from -5.522 (30 samples) to -12.479 (80 samples), suggesting that the model is prone to over-fitting or misled by the limited and possibly biased samples. In contrast, MTL‑FNO displays a consistently ascending or stable trend across all tasks and sample sizes. Even on tasks where FNO eventually performs well—such as the pressure field P in Case A and DSine in Case B—MTL-FNO matches or slightly exceeds the single-field baseline. Overall, these observations demonstrate that multi‑task learning enables MTL‑FNO to effectively extract the shared physical features among coupled fields, thereby serving as a strong regularizer that reduces the dependence on large high‑fidelity datasets. This establishes MTL‑FNO as a reliable surrogate modeling strategy for multi‑field reconstruction under few‑shot conditions.

\section{Conclusion}
\label{sec:conclu}

To achieve an optimal trade-off between prediction accuracy and computational complexity within a multi-field framework, this paper first proposes a lightweight Multi-task Fourier Neural Operator (MTL‑FNO), an end‑to‑end joint training framework with layer-wise parameter sharing. In each MTL-FNO layer, model parameters are partitioned to shared and task-specific components to exploit common latent features across fields and maintain the task-specific characteristics. Furthermore, task-specific fine-tuning parameters are constructed via CP tensor decomposition to reduce the multi-task model size. Second, to address the challenge of co-optimizing and aggregating the parameters in the complex-valued domain, we revisit FNO's spectral convolution from a polar-form perspective, and design a physically meaningful decoupled optimization paradigm. Specially, polar decomposition is employed to slice-wise disentangle the complex-valued weight into a unitary tensor encoding phase information and a positive semi-definite tensor characterizing amplitude. By decoupling the optimization of phase and amplitude, our method can effectively mitigate the task conflicts. Meanwhile, the Cayley transform is introduced to preserve unitary geometric fidelity during training, thereby transforming the constrained optimization on the unitary manifold into an unconstrained problem. Finally, extensive evaluations on two representative engineering benchmarks demonstrate that MTL‑FNO achieves significant advantages in both model compactness and reconstruction accuracy, confirming the effectiveness of its core components—layer-wise parameter partitioning, polar decomposition, and the Cayley transform—and showing that MTL-FNO consistently maintains high accuracy even with very limited training samples. The current work assumes steady-state physical fields and is restricted to two dimensions. Future research will focus on extending MTL‑FNO to high-dimensional, time-dependent physical field reconstruction and prediction, thereby enhancing its applicability and generalization capability in modeling more complex dynamical systems.

\section*{CRediT authorship contribution statement}	
Siyu Ye: Writing – original draft, validation, software, 
methodology, formal analysis.	
Shihang Li: Writing - review and editing, formal analysis.
Zhiqiang Gong: Writing – review and editing, supervision, resources.
Benrong Zhang: Data curation.
Weien Zhou: Supervision, project administration.
Yiyong Huang: Supervision, project administration.
Wen Yao: Supervision, project administration, funding acquisition.

\section*{Declaration of competing interest}
The authors declare that they have no known competing financial interests or personal relationships that could have appeared to influence the work reported in this paper.

\section*{Acknowledgments}
This work was partly supported by the Young Elite Scientist Sponsorship Program of CAST (Grant No. YESS20240697) and the National Natural Science Foundation of China (Grant No. 92371206). The authors also sincerely thank the team of Wenwen Zhao at Zhejiang University, China, for providing the rarefied flow data.

\clearpage
\bibliographystyle{unsrt} 
\bibliography{references}

\begin{thebibliography}{10}

\bibitem{Wang2025Spacecraft}
Kunpeng Wang, Wenhao Yang, Wenbo Li, Yi~Chai, Juan Yao, Xiaofeng Huang, and
  Tong Wang.
\newblock Progress in autonomous intelligent maintenance technologies for
  spacecraft.
\newblock {\em Yuhang Xuebao/Journal of Astronautics}, 46(2):215--231, 2025.

\bibitem{fukami2021gfr}
Kai Fukami, Romit Maulik, Nesar Ramachandra, Koji Fukagata, and Kunihiko Taira.
\newblock Global field reconstruction from sparse sensors with voronoi
  tessellation-assisted deep learning.
\newblock {\em Nature Machine Intelligence}, 3(11):1--7, 2021.

\bibitem{Sharma2018Heat}
Rishi Sharma, Amir~Barati Farimani, Joe Gomes, Peter Eastman, and Vijay Pande.
\newblock Weakly-supervised deep learning of heat transport via physics
  informed loss.
\newblock {\em Statistics}, 2018.

\bibitem{Marco2022GNN}
Marco Maurizi, Chao Gao, and Filippo Berto.
\newblock Predicting stress, strain and deformation fields in materials and
  structures with graph neural networks.
\newblock {\em Scientific Reports}, 12(1):21834, 2022.

\bibitem{Manoj2023onboard}
S.~Manoj, Sannidhi Kasturi, Chandrakala~G. Raju, H.~N. Suma, and Jayanthi~K.
  Murthy.
\newblock Overview of on-board computing subsystem.
\newblock In {\em Smart Small Satellites: Design, Modelling and Development},
  2023.

\bibitem{Lorenzo2024Onboard}
Lorenzo Diana and Pierpaolo Dini.
\newblock Review on hardware devices and software techniques enabling neural
  network inference onboard satellites.
\newblock {\em Remote Sensing}, 16(21):3957, 2024.

\bibitem{Olaf2015U-Net}
Olaf Ronneberger, Philipp Fischer, and Thomas Brox.
\newblock {U-Net}: Convolutional networks for biomedical image segmentation.
\newblock In {\em Medical Image Computing and Computer-Assisted Intervention --
  MICCAI 2015}, 2015.

\bibitem{Zhao2023cnn}
Xiaoyu Zhao, Zhiqiang Gong, Yunyang Zhang, Wen Yao, and Xiaoqian Chen.
\newblock Physics-informed convolutional neural networks for temperature field
  prediction of heat source layout without labeled data.
\newblock {\em Engineering Applications of Artificial Intelligence}, 117(Part
  A):105516, 2023.

\bibitem{Jordan2024U-Net}
Jordan A.~C. Kildare, Wai~Tong Chung, Michael~J. Evans, Zhao~F. Tian, Paul~R.
  Medwell, and Matthias Ihme.
\newblock Predictions of instantaneous temperature fields in jet-in-hot-coflow
  flames using a multi-scale {U-Net} model.
\newblock {\em Proceedings of the Combustion Institute}, 40(1):105330, 2024.

\bibitem{Tong2025PIU-Net}
Tong Zhu, Dehao Liu, and Yanglong Lu.
\newblock Finite-volume physics-informed {U-Net} for flow field reconstruction
  with sparse data.
\newblock {\em Journal of Computing and Information Science in Engineering},
  25(7):071004, 2025.

\bibitem{Lu2021DeepONet}
Lu~Lu, Pengzhan Jin, Guofei Pang, Zhongqiang Zhang, and George~Em Karniadakis.
\newblock Learning nonlinear operators via {DeepONet} based on the universal
  approximation theorem of operators.
\newblock {\em Nature Machine Intelligence}, 3(3):218--229, 2021.

\bibitem{Li2021FNO}
Zongyi Li, Nikola Kovachki, Kamyar Azizzadenesheli, Burigede Liu, Kaushik
  Bhattacharya, Andrew Stuart, and Anima Anandkumar.
\newblock Fourier neural operator for parametric partial differential
  equations.
\newblock 2021.

\bibitem{Santos2023Senseiver}
Javier~E. Santos, Zachary~R. Fox, Arvind Mohan, Daniel O'Malley, Hari
  Viswanathan, and Nicholas Lubbers.
\newblock Development of the senseiver for efficient field reconstruction from
  sparse observations.
\newblock {\em Nature Machine Intelligence}, 5(11):1317--1325, 2023.

\bibitem{Ma2025PhySense}
Yuezhou Ma, Haixu Wu, Hang Zhou, Huikun Weng, Jianmin Wang, and Mingsheng Long.
\newblock {PhySense}: Sensor placement optimization for accurate physics
  sensing.
\newblock {\em arXiv preprint}, 2025.

\bibitem{Zhu2025pruning}
Kehan Zhu, Fuyi Hu, Yuanbing Ding, Wei Zhou, and Ruxin Wang.
\newblock A comprehensive review of network pruning based on pruning
  granularity and pruning time perspectives.
\newblock {\em Neurocomputing}, page 129382, 2025.

\bibitem{Babak202Quantization}
Babak Rokh, Ali Azarpeyvand, and Alireza Khanteymoori.
\newblock A comprehensive survey on model quantization for deep neural networks
  in image classification.
\newblock {\em ACM Transactions on Intelligent Systems and Technology},
  14(6):1--50, 2023.

\bibitem{Amir2025Distillation}
Amir~M. Mansourian, Rozhan Ahmadi, Masoud Ghafouri, Amir~Mohammad Babaei,
  Elaheh Badali~Golezani, Zeynab Yasamani~Ghamchi, Vida Ramezanian, Alireza
  Taherian, Kimia Dinashi, Amirali Miri, and Shohreh Kasaei.
\newblock A comprehensive survey on knowledge distillation.
\newblock 2025.

\bibitem{OU2024lowrank}
Xinwei Ou, Zhangxin Chen, Ce~Zhu, and Yipeng Liu.
\newblock Low rank optimization for efficient deep learning: Making a balance
  between compact architecture and fast training.
\newblock {\em Journal of Systems Engineering and Electronics}, 35(3):509--531,
  2024.

\bibitem{Jon2025On1D-CNN}
Jon Alvarez~Justo, Dennis~D. Langer, Simen Berg, Jens Nieke, Radu~Tudor
  Ionescu, Per~Gunnar Kjeldsberg, and Tor~Arne Johansen.
\newblock Hyperspectral image segmentation for optimal satellite operations:
  In-orbit deployment of {1D-CNN}.
\newblock {\em Remote Sensing}, 17(4):642, 2025.

\bibitem{Roberto2025onnas}
Roberto Del~Prete, Parampuneet~Kaur Thind, Andrea Mazzeo, Matthew Whitley,
  Lorenzo Papa, Nicolas Long{\'e}p{\'e}, and Gabriele Meoni.
\newblock Optimizing deep learning models for on-orbit deployment through
  neural architecture search.
\newblock {\em Scientific Reports}, 15(1):37783, 2025.

\bibitem{Dimitrios2025Edge}
Dimitrios Meimetis, Ioannis Daramouskas, Niki Patrinopoulou, Vaios Lappas, and
  Vassilis Kostopoulos.
\newblock Comparative analysis of object detection models for edge devices in
  {UAV} swarms.
\newblock {\em Machines}, 13(8):684, 2025.

\bibitem{Bharadwaj2026onboard}
Bharadwaj Chintalapati, Aras Jafari, Rene Laufer, Marcus Liwicki, and Jens
  Eickhoff.
\newblock Advancing multi-class object detection from {LEO/VLEO}: Model
  evaluation and onboard deployment tailored for a {16U} {CubeSat}.
\newblock {\em CEAS Space Journal}, 2026.

\bibitem{Sarthak2022Comparison}
Sarthak Kapoor, Jaber~Rezaei Mianroodi, Mohammad Khorrami, Nima~S. Siboni, and
  Bob Svendsen.
\newblock Comparison of two artificial neural networks trained for the
  surrogate modeling of stress in materially heterogeneous elastoplastic
  solids.
\newblock 2022.

\bibitem{Jinghong2025lamdaFNO}
Jinghong Xu, Yuqian Zhou, Qian Liu, Kebing Li, and Haolin Yang.
\newblock Transfer learning of neural operators for partial differential
  equations based on sparse network $\lambda$-{FNO}.
\newblock {\em {PLoS ONE}}, 20(5):e0321154, 2025.

\bibitem{Yu2024TransferFNO}
Yufeng Wang, Heng Zhang, Chensen Lai, and Xiangyun Hu.
\newblock Transfer learning fourier neural operator for solving parametric
  frequency-domain wave equations.
\newblock {\em IEEE Transactions on Geoscience and Remote Sensing}, pages
  1--11, 2024.

\bibitem{Brunton2021Data-Driven}
Steven~L. Brunton, J.~Nathan Kutz, Krithika Manohar, Aleksandr~Y. Aravkin,
  Kristi Morgansen, Jennifer Klemisch, Nicholas Goebel, James Buttrick, Jeffrey
  Poskin, Adriana~W. Blom-Schieber, Thomas Hogan, and Darren McDonald.
\newblock Data-driven aerospace engineering: Reframing the industry with
  machine learning.
\newblock {\em AIAA Journal}, 59(8):2820--2847, 2021.

\bibitem{Ma2025multisource}
Xiaobing Ma, Xuyi Jia, Chunna Li, and Chunlin Gong.
\newblock A sparse reconstruction method of physical field via multi-source
  sensors for flight vehicle.
\newblock {\em Aerospace Science and Technology}, page 110685, 2025.

\bibitem{Maxime2024Multitask}
Maxime Fontana, Michael Spratling, and Miaojing Shi.
\newblock When multitask learning meets partial supervision: A computer vision
  review.
\newblock {\em Proceedings of the IEEE}, 112(6):516--543, 2024.

\bibitem{Wang2023AdaptiveHPS}
Hongxia Wang, Xiao Jin, Yukun Du, Nan Zhang, and Hongxia Hao.
\newblock Adaptive hard parameter sharing method based on multi-task deep
  learning.
\newblock {\em Mathematics}, 11(22):4639, 2023.

\bibitem{Graham2023Mtl}
Simon Graham, Quoc~Dang Vu, Mostafa Jahanifar, Shan E.~Ahmed Raza, Fayyaz
  Minhas, David Snead, and Nasir Rajpoot.
\newblock One model is all you need: Multi-task learning enables simultaneous
  histology image segmentation and classification.
\newblock {\em Medical Image Analysis}, 83(Suppl C):102685, 2023.

\bibitem{Zhang2023MTL}
Yi~Zhang, Yu~Zhang, and Wei Wang.
\newblock Learning linear and nonlinear low-rank structure in multi-task
  learning.
\newblock {\em IEEE Transactions on Knowledge and Data Engineering},
  (8):8157--8170, 2023.

\bibitem{Cheng2025Multitask}
Dianjing Cheng, Xiangyu Shi, Zhihua Cui, Xingyu Wu, and Wenjia Niu.
\newblock Multitask vehicle signal recognition with dual-speed adaptive
  weighting.
\newblock {\em Journal of Advanced Transportation}, 35(1):9961530, 2025.

\bibitem{garg2025ftn}
Yash Garg, Nebiyou Yismaw, Rakib Hyder, Ashley Prater-Bennette, Amit
  Roy-Chowdhury, and M.~Salman Asif.
\newblock Parameter-efficient multi-task and multi-domain learning using
  factorized tensor networks.
\newblock {\em IEEE Open Journal of Signal Processing}, pages 1077--1085, 2025.

\bibitem{Wang2023PEM}
Haowen Wang, Tao Sun, Cong Fan, and Jinjie Gu.
\newblock Customizable combination of parameter-efficient modules for
  multi-task learning.
\newblock 2023.

\bibitem{Zhang2025MoRE}
Dacao Zhang, Kun Zhang, Shimao Chu, Le~Wu, Xin Li, and Si~Wei.
\newblock {MoRE}: A mixture of low-rank experts for adaptive multi-task
  learning.
\newblock 2025.

\bibitem{He2025RaSA}
Zhiwei He, Zhaopeng Tu, Xing Wang, Xingyu Chen, Zhijie Wang, Jiahao Xu, Tian
  Liang, Wenxiang Jiao, Zhuosheng Zhang, and Rui Wang.
\newblock {RaSA}: Rank-sharing low-rank adaptation.
\newblock 2025.

\bibitem{Yang2024MTL-LoRA}
Yaming Yang, Dilxat Muhtar, Yelong Shen, Yuefeng Zhan, Jianfeng Liu, Yujing
  Wang, Hao Sun, Denvy Deng, Feng Sun, Qi~Zhang, Weizhu Chen, and Yunhai Tong.
\newblock {MTL-LoRA}: Low-rank adaptation for multi-task learning.
\newblock 2024.

\bibitem{Senushkin2023Alignment}
Dmitry Senushkin, Nikolay Patakin, Arseny Kuznetsov, and Anton Konushin.
\newblock Independent component alignment for multi-task learning.
\newblock In {\em 2023 IEEE/CVF Conference on Computer Vision and Pattern
  Recognition (CVPR)}, 2023.

\bibitem{Chen2023benchmark}
Xiaoqian Chen, Zhiqiang Gong, Xiaoyu Zhao, Weien Zhou, and Wen Yao.
\newblock A machine learning surrogate modeling benchmark for temperature field
  reconstruction of heat source systems.
\newblock {\em Science China Information Sciences}, 66(5):152203, 2023.

\end{thebibliography}

\appendix
\section{Physical Conditions for High-Fidelity Simulation of Two-Dimensional Rarefied Flow over a Blunt Wedge}
\label{sec:append}
\begin{figure}[htbp]
	\centering
	\includegraphics[width=0.4\linewidth]{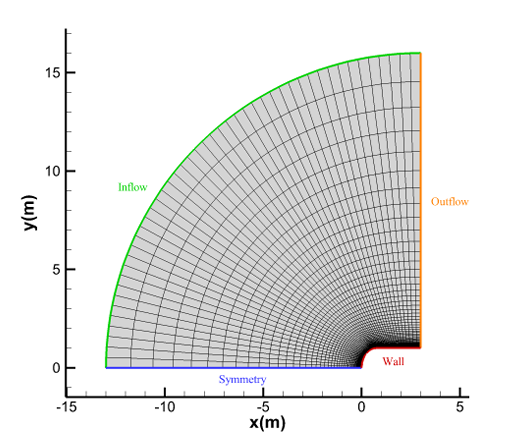}
	\caption{Computational mesh and boundary conditions.}
	\label{fig:dunxiewangge}
\end{figure}

The physical setup, computational mesh, and boundary conditions for the high-fidelity numerical simulation of the two-dimensional blunt wedge are illustrated in Fig.~\ref{fig:dunxiewangge} and summarized as follows:
\begin{itemize}[leftmargin=*, nosep]
	\item Altitude: $100\,\mathrm{km}$
	\item Freestream Mach number: $Ma = 5$
	\item Freestream pressure: $p_\infty = 0.0320113\,\mathrm{Pa}$
	\item Freestream temperature: $T_\infty = 195.081\,\mathrm{K}$
	\item Working gas: Monatomic argon (Ar)
	\item Dynamic viscosity: $\mu = 5.071158 \times 10^{-5}\,\mathrm{Pa \cdot s}$
	\item Specific heat ratio: $\gamma = 1.667$
	\item Prandtl number: $Pr = 0.667$
	\item Specific gas constant: $R = 208.16\,\mathrm{J/(kg \cdot K)}$
	\item Inverse power-law exponent: $\omega = 0.734$
\end{itemize}

\end{document}